%% file: main.tex
\documentclass{article}
\usepackage[utf8]{inputenc}
\usepackage{apacite}
\usepackage{graphicx}
\usepackage{amsfonts}
\usepackage{color}
\usepackage{amsthm}
\usepackage{amssymb}
\usepackage{enumitem}
\usepackage[mathscr]{euscript}
 \let\mathscr\relax% just so we can load this and rsfs
\usepackage[scr]{rsfso}
\usepackage{comment}
\usepackage{amsmath}

\newcommand{\scc}[1]{\textcolor{red}{SC: #1}}

\newcommand{\pmap}{\multimap\!\rightarrow}
\newcommand{\power}[1]{\mathscr{P}(#1)}

\newcommand{\mf}[1]{\mathsf{#1}}

\newtheorem{definition}{Definition}
\newtheorem{proposition}{Proposition}

\title{Formalising Concepts as Grounded Abstractions}

\author{Stephen Clark, Alexander Lerchner\and Tamara von Glehn, Olivier Tieleman, Richard Tanburn\and Misha Dashevskiy, Matko Bo\v{s}njak\\ DeepMind, London, UK}

\date{January 2021}

\begin{document}

\maketitle

\include{introduction}
\include{discrete}
\include{instances}

\include{probabilities}
\include{conclusion}

\section{Acknowledgements}

Thanks to Richard Evans and the DeepMind Concepts team for useful feedback.

\bibliographystyle{apacite}
\bibliography{references}

\end{document}

%% file: introduction.tex
\section{Introduction}

The notion of \emph{concept} has been studied for centuries, by philosophers, linguists, cognitive scientists, and researchers in artificial intelligence \cite{margolis_core_readings}. There is a large literature on formal, mathematical models of concepts, including a whole sub-field of AI---Formal Concept Analysis---devoted to this topic \cite{Ganter2016}. Recently, researchers in machine learning have begun to investigate how methods from representation learning can be used to induce concepts from raw perceptual data \shortcite{SCAN}. The goal of this report is to provide a formal account of concepts which is compatible with this latest work in deep learning. 

Since the concepts literature is so large, and covers so many disciplines, we will not attempt to survey the whole field, but rather provide links to the parts of the literature which are especially relevant to our own work. Good places to start for an introduction to concepts include \citeA{margolis_core_readings,conceptual_mind,stanford_encyc}, \citeA{murphy_concepts}, and \citeA{gardenfors2014}. 

The main technical goal of this report is to show how techniques from representation learning can be married with a lattice-theoretic formulation of conceptual spaces. The mathematics of partial orders and lattices is a standard tool for modelling conceptual spaces (Ch.2, \citeA{mitchell_ml}, \citeA{Ganter2016}); however, there is no formal work that we are aware of which defines a conceptual lattice on top of a representation that is induced using unsupervised deep learning \shortcite{goodfellow_dl}. \citeA{SCAN} do this to a degree, but here we provide a much more comprehensive and formal account. \citeA{gardenfors} offers a geometric account which fits naturally with representation learning, and we will be drawing some inspiration from G\"{a}rdenfors' work, but with more of a focus on how concepts can be ordered. The advantages of partially-ordered lattice structures are that these provide natural mechanisms for use in concept discovery algorithms, through the meets and joins of the lattice. Finally, although we do not provide much background in terms of the concepts literature, we do attempt some rigour in the mathematical presentation of lattices and partial orders, which will be based heavily on \citeA{lattices_2nd_ed}.

Overall, our aim is to provide a formal framework for developing practical conceptual discovery and reasoning systems which are \emph{grounded} in perception and action \cite{harnad}, thereby overcoming a fundamental deficiency in formal representation systems which are either constructed manually by a knowledge engineer (Ch.8, \citeA{russell_norvig}) or induced automatically from purely text-based resources \shortcite{banko2007}. Note that the main aim of this report is a \emph{mathematical} one; how to realize the framework in practice, and how the framework relates to the vast literature on concepts---across philosophy, psychology, linguistics and AI---are questions left largely for future work.

The rest of the report is organised as follows. Section~\ref{sec:formalisation} introduces the mathematics of partial orders and lattices by first considering the case of discrete feature values. Section~\ref{sec:instances} then introduces representation learning into the picture, by defining the notion of an \emph{instance space}, which is the space that an intelligent agent uses to represent its environment. Instances are the representations over which abstraction takes place, and abstraction is the operation which leads to the conceptual lattice structure. This section will also extend the treatment of finite feature values to the continuous case, showing how grouping values into sets naturally maintains a partial order. Section~\ref{sec:CPOs} provides the main mathematical result of this work, which is that concepts can be defined as elements of a complete partial order (CPO), with instances as limiting cases of concepts (maximal elements of the CPO). Finally, Section~\ref{sec:probs} suggests ways in which probabilities can be introduced into the mix, which is an important question given that the representation learning techniques we consider are inherently probabilistic.

%% file: discrete.tex
\section{Conceptual Spaces as Discrete Feature Lattices}
\label{sec:formalisation}

The \emph{classical theory of concepts} is based on the idea that concepts are essentially definitions, providing necessary and sufficient conditions for membership of the extension of a concept \shortcite{stanford_encyc}. These conditions are often given in terms of defining \emph{features}; a standard example is the concept of \textsc{bachelor} having the features $\mf{human}$, $\mf{unmarried}$ and $\mf{male}$. One of the issues with the classical view is the question of where these features come from. Here we assume that recent developments in representation learning can answer that question, by inducing a representation space with separable dimensions (or more generally separable sub-spaces) which provide the conceptual features \shortcite{beta-vae,higgins:disentangled}. 

We certainly don't want to commit to the classical view in general, in particular because we do not define concepts as definitions. Moreover, there are a number of additional issues with the classical view, which motivate the other main concept theories (the \emph{prototype}, \emph{exemplar}, and \emph{theory} theories \shortcite{stanford_encyc,murphy_concepts}). The defining characteristic of our approach is that concepts are \emph{grounded abstractions} \cite{SCAN}; \emph{grounded} because we assume a mechanism for inducing concepts from perceptual input, and this explains how concepts are related to the external world; and \emph{abstractions} because these allow for efficient, combinatorial planning and reasoning in an intelligent agent \cite{lake_thinking_machines}.

\subsection{Discrete Features and Partial Maps}
\label{sec:discrete_feats}

From the mathematical perspective, we need a  formalism which can represent sets of feature-value pairs, and also the operations which combine two such sets. There are a number of areas of AI which have made extensive use of feature structures, for example knowledge representation (Ch.2, \citeA{mitchell_ml}), formal concept analysis \cite{Ganter2016}, and computational linguistics \cite{Carpenter92}. In all these cases, the underlying mathematical structures are based on partial orders, and more specifically lattices. Theoretical computer science is another area that has made extensive use of such structures \cite{abramsky_domains}.
A useful intuition to start with is the idea that, when combining two concepts, the result should be a) consistent with both of the combining concepts; and b) have no more additional information in it than is already present in those concepts. Any reader familiar with formal frameworks for knowledge representation, or perhaps logic programming, may recognise this as an informal description of \emph{unification} (Ch.9, \citeA{russell_norvig}), which also relies on the mathematics of partial orders.

Much of this section follows the presentation of lattices and partial orders in \citeA{lattices_2nd_ed} (D\&P). Note also that there is nothing particularly new in this section, and our presentation of discrete feature lattices is similar to the many other presentations found in the fields mentioned above. However, the standard discrete case forms a useful basis on which to develop some of the continuous conceptual lattices described later.

To begin with, let's assume a finite set of features $\mathsf{Feat}$ and a finite set of values $\mathsf{Val}$. In Section~\ref{sec:instances} we'll relax the finite values assumption and consider an infinite set of (potentially continuous) values. One practical way to obtain a finite set of values, at least from a (totally) ordered infinite set, is to create equivalence classes using boundary cutoffs. For example, suppose we have some feature which has values in $\mathbb{R}$, then we can create a finite number of values by partitioning the real number line into ``buckets"; e.g. bucket 1 has all values less than some threshold $x_1$, bucket 2 all values in $[x_1,x_2)$, and bucket $N$ all values greater than or equal to $x_N$, where $x_i < x_k,$ for $i < k$ . 

To make the discussion more concrete, consider the following set of features: $\mathsf{Feat} = \{ \mathsf{Color, Shape, Weight, Position} \}$. Each feature has an associated set of possible values, say:\\

\begin{tabular}{lll}
$\mf{Val}(\mf{Color})$ & = & $\{ \mf{Red, Blue, Green, Black, White} \}$,\\
$\mf{Val}(\mf{Shape})$ & = & $\{ \mf{Circle, Square, Triangle, Diamond} \}$,\\
$\mf{Val}(\mf{Weight})$ & = & $\{ \mf{Light, Medium, Heavy} \}$,\\
$\mf{Val}(\mf{Position})$ & = & $\{ \mf{Center, TopLeft, TopRight, BottomLeft, BottomRight} \}$. 
\end{tabular}\\

\noindent
We'll denote the union of all the feature values as $\mf{Val}$.

The key aspect of an abstraction is that it is missing some information. Phrases in D\&P used to express this notion include \emph{greater} or \emph{lesser information content}, and being \emph{more} or \emph{less informative than}. One concept, or abstraction, that we can form from the features above is \textsc{Cannonball}, which we might define as a heavy, black circle. Since \textsc{Cannonball} has 3 out of a possible 4 features defined, this would be relatively informative or have high information content.

Mathematically, the way to express an abstraction based on these intuitions is with a \textbf{partial map} (or \textbf{partial function}).\footnote{All mathematical definitions in this section, and some of the accompanying text (in quotes when taken verbatim), are from D\&P.} Let $X$ and $Y$ be non-empty sets and $f : X \rightarrow Y$ a map; $f$ can be thought of as a process which assigns a member $f(x)$ of $Y$ to each $x\in X$, or equivalently as a \textbf{graph}, i.e. the set of argument-value pairs defining the map: $\{ (x,f(x))\,\vert\, x\in X \}$.

``If the values of $f$ are given on some subset $S$ of $X$, we have partial information towards determining $f$."

\begin{definition}
A \emph{partial map} from $X$ to $Y$ is a map $\sigma : S \rightarrow Y$, where \emph{dom}$\,\sigma$, the domain of $\sigma$, is a subset $S$ of $X$ ($S$ can be $\emptyset$). \emph{(p.7, D\&P)}
\end{definition}

\noindent
If dom$\,\sigma = X$, then $\sigma$ is a map (or a \textbf{total map}) from $X$ to $Y$. The set of partial maps (which includes the total maps) from $X$ to $Y$ is denoted $(X \pmap Y)$.

Now we are in a position to define a concept as an  abstraction:\footnote{There is a technicality here in that the range of the partial map is partitioned into sets corresponding to different values of the domain (e.g. the value of the $\mf{Color}$ feature cannot be $\mf{TopLeft}$), but we'll gloss over that for now.}

\begin{definition}
\label{def:concept}
Assuming a finite set of features $\mf{Feat}$ and a finite set of values $\mf{Val}$, a concept $C$ is a partial map from $\mf{Feat}$ to $\mf{Val}$, i.e. a map $\delta_C : \mf{SubFeat} \rightarrow \mf{Val}$, where $\mf{SubFeat} \subseteq \mf{Feat}.$
\end{definition}

%\tvg{Since a partial map has just been defined and SubFeat is not used again, would it be clearer to define a concept as a partial map $\mf{Feat} \rightarrow \mf{Val}$?}

\noindent
Equivalently, each concept is a set of feature-value pairs (the corresponding graph). Continuing the earlier example, the concept of a cannonball would be defined as:
\[ \textsc{Cannonball} = \{ \langle \mf{Color,Black}\rangle, \langle \mf{Shape,Circle} \rangle, \langle \mf{Weight, Heavy} \rangle \}. \] 
\noindent
Notice this is an abstraction (partial map) over $\mf{Feat}$ since $\mf{Position}$ has no value.

What sorts of binary combination operations might we want to perform over these sets of feature-value pairs? The obvious ones are set union, intersection and difference. For example:\\

\begin{tabular}{l}
$\{ \langle \mf{Color,Red}\rangle \}\, \cup \, \{ \langle \mf{Shape,Circle}\rangle \}$  =  $\{ \langle \mf{Color,Red}\rangle, \langle \mf{Shape,Circle}\rangle \}$\\
$\{ \langle \mf{Color,Black}\rangle, \langle \mf{Shape,Circle}\rangle \}\, \cap \, \{ \langle \mf{Shape, Circle}\rangle, \langle \mf{Weight, Heavy}\rangle \}$  =\\
$\;\;\;\{\langle \mf{Shape,Circle}\rangle \}$\\
$\{ \langle \mf{Color,Black}\rangle, \langle \mf{Shape,Circle}\rangle \}\, \setminus \{ \langle \mf{Color,Black}\rangle \} = \{ \langle \mf{Shape,Circle}\rangle \}$\\
\end{tabular}\\

The intuition behind the first example is that, given the concepts \textsc{Red} and \textsc{Circle}, we can, through the set union operation, form a new concept \textsc{RedCircle}. Similarly, for the second example, from \textsc{BlackCircle} and \textsc{HeavyCircle}, and the application of set intersection, we can form \textsc{Circle}. And finally, from \textsc{BlackCircle} and \textsc{Black}, and the application of set difference, we can form \textsc{Circle}.

\subsection{Partial Orders and Bounds}

What is the algebraic structure underlying these operations? Set union and intersection applied to the elements of the power set of some set provides a textbook example of a \emph{lattice}, which relies on the notion of a \emph{partial order}.

\begin{definition}
Let $P$ be a set. An \textbf{order} or \textbf{partial order} on $P$ is a binary relation $\leq$ on $P$ such that, for all $x,y,z \in P$,
\begin{enumerate}[label=(\roman*)]
    \item $x \leq x$,
    \item $x \leq y$ and $y \leq x$ imply $x = y$,
    \item $x \leq y$ and $y \leq z$ imply $x \leq z$. \emph{(p.2, D\&P)}
\end{enumerate}
\end{definition}

\noindent
These conditions define a partial order as a relation that is i) reflexive, ii) antisymmetric, and iii) transitive. A set $P$ equipped with an order relation $\leq$ is said to be a (partially) ordered set, or \emph{poset}.   

There are other types of order resulting from different sets of constraints. For example, a relation $\leq$ on a set $P$ which is reflexive and transitive but not necessarily antisymmetric is a \emph{quasi-order} or \emph{pre-order}. A partially ordered set which has the additional condition that, for all $x,y \in P$, either $x \leq y$ or $y \leq x$ (i.e. any two elements of $P$ are comparable), is called a \emph{chain} or \emph{linearly ordered set} or \emph{totally ordered set}. If two elements $x$ and $y$ are not comparable in the order, i.e. $x \nleq y$ and $y \nleq x$, then we write $x\,\|\,y$.

\begin{definition}
\textbf{Bottom and top.} Let $P$ be an ordered set. We say $P$ has a bottom element if there exists $\bot \in P$ (called \textbf{bottom}) with the property that $\bot \leq x$ for all $x \in P$. Dually, $P$ has a top element if there exists $\top \in P$ such that $x \leq \top$ for all $x \in P$. \emph{(p.15, D\&P)}
\end{definition}

Consider a set $P = \mathscr{P}(X)$, the powerset of some set $X$, then $\langle P;\subseteq \rangle$ is a partial order. It is easy to check that the subset relation is reflexive, antisymmetric, and transitive. We also have that in $\langle P;\subseteq \rangle$, $\bot = \emptyset$ and $\top = X$.

In order to progress to lattices, we need the notions of least upper bound and greatest lower bound. Here is how these are defined in D\&P:

\begin{definition}
Let $P$ be an ordered set and let $S \subseteq P$. An element $x \in P$ is an \textbf{upper bound} of $S$ if $s \leq x$ for all $s \in S$. A \textbf{lower bound} is defined dually. The set of all upper bounds of $S$ is denoted by $S^u$ (read as `S upper') and the set of all lower bounds by $S^l$ (read as `S lower'):
\[ S^u = \{ x\in P \,\vert\, (\forall s \in S) \,s \leq x \} \,\mbox{ \emph{and} }\, S^l = \{ x\in P \,\vert\, (\forall s \in S) \,s \geq x \}\]
\end{definition}

``If $S^u$ has a least element $x$, then $x$ is called the \textbf{least upper bound} of $S$. Equivalently, $x$ is the least upper bound of $S$ if
\begin{enumerate}[label=(\roman*)]
    \item $x$ is an upper bound of $S$, and
    \item $x \leq y$ for all upper bounds $y$ of $S$.
\end{enumerate}

Dually, if $S^l$ has a greatest element, $x$, then $x$ is called the \textbf{greatest lower bound} of $S$. Since least elements and greatest elements are unique, least upper bounds and greatest lower bounds are unique when they exist." (p.33, D\&P)

There are some alternative notations and terminology for referring to bounds. 
The least upper bound of $S$ is also called the \textbf{supremum} of $S$ and is denoted by $\mbox{sup}\,S$; the greatest lower bound of $S$ is also called the \textbf{infimum} of $S$ and is denoted $\mbox{inf}\,S$. The  notation and terminology we will use in the rest of the report is based on the terms \emph{meet} and \emph{join}. We will write $x \vee y$ (`$x$ \textbf{join} $y$') instead of sup$\{ x,y \}$ and $x \wedge y$ (`$x$ \textbf{meet} $y$') instead of inf$\{ x,y \}$. We can also write $\bigvee S$ (the `\textbf{join} of $S$') and $\bigwedge S$ (the `\textbf{meet} of $S$') instead of $\mbox{sup}\,S$ and $\mbox{inf}\,S$.

\subsection{Lattices}

Lattices are particular cases of ordered sets in which $x \vee y$ and $x \wedge y$ exist for all $x,y \in P$. These are the mathematical structures that will form the basis of all the conceptual feature lattices described in the rest of the report.

\begin{definition}
Let $P$ be a non-empty ordered set.
\begin{enumerate}[label=(\roman*)]
    \item If $x \vee y$ and $x \wedge y$ exist for all $x,y \in P$, then $P$ is called a \textbf{lattice}.
    \item If $\bigvee S$ and $\bigwedge S$ exist for all $S \subseteq P$, then $P$ is called a \textbf{complete lattice}. \hspace*{\fill}\emph{(p.34, D\&P)}
\end{enumerate}
\end{definition}

As an example, for any set $X$, and some indexing set $I$ which picks out elements of $\mathscr{P}(X)$, the poset $\langle \mathscr{P}(X) ; \subseteq \rangle$ is a complete lattice in which

\[ \bigvee\{ A_i \,\vert\, i \in I \} = \bigcup \,\{ A_i \,\vert\, i \in I \} \]
\[ \bigwedge\{ A_i \,\vert\, i \in I \} = \bigcap \,\{ A_i \,\vert\, i \in I \}. \]

\begin{figure}
\begin{center}
     \includegraphics[width=3cm]{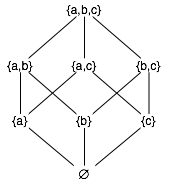}
    \caption{Powerset lattice with the subset relation.}
    \label{fig:powerset_lattice}
\end{center}
\end{figure}

\noindent
Figure~\ref{fig:powerset_lattice} shows a diagram of the subset relation, in the form of what is known as a \textbf{Hasse diagram}. The lines represent the \textbf{covering relation} underlying the partial order, with the ordering going upwards on the page, and with implied transitivity.\footnote{We could orientate the diagram in the other direction, but having the lesser elements at the bottom is consistent with D\&P, and fits with the intuition that lesser in the ordering corresponds to lower on the page. Another way to reverse the orientation is to use the dual relation, in this case the superset relation.} So the fact that there is a line from $\{ b \}$ to $\{ a,b\}$, for example (since $\{ b \} \subseteq \{ a,b\}$) and from $\{ a,b \}$ to $\{ a,b,c\}$ (since $\{ a,b \} \subseteq \{ a,b,c\}$), implies that $\{ b \} \subseteq \{ a,b,c\}$.

It's easy to see from the diagram why $\langle \mathscr{P}(X) ; \subseteq \rangle$ is a lattice: take any pair in the diagram, and follow the lines upward from each element of the pair. In all cases, the lines will intersect, and where there is more than one intersection point, one of those points will be lower than the other intersection points, which is the least upper bound of the pair (or the join). In the most extreme case, the join will be at the top ($\top = \{ a,b,c \}$). A similar comment applies to following lines downward, in which case two of the lines will intersect at the greatest lower bound, or the meet -- in the most extreme case at the bottom ($\bot = \emptyset$).

A lattice with a top $\top$ and bottom $\bot$ element is called \textbf{bounded}. A finite lattice $L$ is automatically bounded, with $\top = \bigvee L$ and $\bot = \bigwedge L$. All finite lattices are also complete.

\subsubsection{Set Difference}
\label{sec:set_diff}

We might also like to apply set difference (also known as set minus) to two concepts (as we did in the $\textsc{BlackCircle}$ example above), and hence give a definition of set difference in terms of partial orders. An initial thought might be that we can define set difference in terms of union and intersection, but that is not possible: we also need set complement. D\&P do define complements in lattices, but do not deal with the case of set difference. The definition of complements is as follows.

\begin{definition}
\textbf{Complements}. Let $L$ be a bounded lattice with $\bot$ and $\top$ elements. For $x \in L$, we say $z$ is a \textbf{complement} of $x$ if $x \wedge z = \bot$ and $x \vee z = \top$. Note that complements are not necessarily unique, and some elements may have no complement.
\end{definition}

To get to set difference we need to extend the definition to \emph{relative complements}, which we can do as follows \shortcite{math_methods_ling}.

\begin{definition}
\textbf{Relative complements}. Let $L$ be a bounded lattice with $\bot$ and $\top$ elements. For $x,y \in L$, where $x\leq y$, we say $z$ is a \textbf{complement} of $x$ \textbf{relative to} $y$ if $z$ is a complement of $x$ in the sub-partial order below $y$, i.e.
$x \wedge z = \bot$ and $x \vee z = y$.
\end{definition}

Now consider the lattice $\langle \mathscr{P}(U) ; \subseteq \rangle$, for some set $U$ (an example of which is in Figure~\ref{fig:powerset_lattice}), and the set difference $Y - X$ for $X,Y \in \mathscr{P}(U)$ and $X \subseteq Y$. The complement of $X$ relative to $Y$ is the set difference in this case:

\[ Y \setminus X = \{x \in U \;|\; x \in Y, x \notin X\} \]

\noindent
Note that $X \cap (Y \setminus X) = \emptyset$ and $X \cup (Y \setminus X) = Y$.

In the more general case, we may not have $x \leq y$ (or $X \subseteq Y$). For example, $\{ a,b\} \setminus \{b,c\} = \{a\}$ according to the definition of set difference above. In this case, we take the complement of $y \wedge x$ relative to $y$. So for $\{ a,b\} \setminus \{b,c\}$, we take the complement of $\{ a,b\} \cap \{b,c\} = \{b\}$, relative to $\{a,b\}$, which is $\{a\}$.

%\tvg{I think it would make sense to define set difference in terms of `relative complements'. A complement of an element $x$ in a bounded partial order is a (not necessarily unique) $z$ such that $x \vee z = \top$ and $x \wedge z = \bot$. If $x \leq y$, a complement of $x$ relative to $y$ is a complement in the sub-partial order below $y$, i.e. an element $z$ such that $x \vee z = y$ and $x \wedge z = \bot$. In $\mathscr{P}(X)$ this gives exactly the set difference}
%$$S \setminus T = \{x \in X \;|\; x \in S, x \notin T\}$$
%(which always exists and is unique, i.e. $\mathscr{P}(X)$ is a Boolean lattice). There are also weaker versions of `relative pseudocomplements' if complements don't all exist but the exact definitions seem to vary. I think this version should be enough for concepts.}

\subsection{Orders on Partial Maps}
\label{sec:orders_partial}

Let's go back to the concept example from earlier, and, in order that we can easily draw a diagram, we'll reduce the number of values for each feature:\\

\begin{tabular}{lll}
$\mf{Val}(\mf{Color})$ & = & $\{ \mf{Red, Blue} \}$,\\
$\mf{Val}(\mf{Shape})$ & = & $\{ \mf{Circle, Square} \}$,\\
$\mf{Val}(\mf{Weight})$ & = & $\{ \mf{Light, Heavy} \}$,\\
$\mf{Val}(\mf{Position})$ & = & $\{ \mf{Top, Bottom} \}$. 
\end{tabular}\\

We would like to order the concepts to produce a diagram like that in Figure~\ref{fig:powerset_lattice}, which essentially tells us how to combine two concepts (by finding either the meet or the join). But there's a problem: what do we do in the case where the values of a feature clash? For example, what is the order between the concepts \textsc{Cannonball} and \textsc{RedBalloon}:
\[ \{ \langle\mf{Color,Black}\rangle, \langle \mf{Shape,Circle} \rangle, \langle \mf{Weight, Heavy}\rangle\} \] 
\[ \{\langle\mf{Color,Red}\rangle, \langle \mf{Shape,Circle} \rangle, \langle \mf{Weight, Light}\rangle\} \,? \]

D\&P provide the following order, which neatly captures the idea that, not only do we want to take set unions (and intersections) of the feature-value pairs, but we also need the combining concepts (partial maps) to be compatible.

\begin{definition}
\label{def:pmap_order}
The set of partial maps $(X\pmap Y)$ is ordered as follows. Given $\sigma,\tau \in (X\pmap Y)$, define $\sigma \leq \tau$ if and only if $\mbox{\emph{dom}}\,\sigma \subseteq \mbox{\emph{dom}}\,\tau$ and $\sigma(x) = \tau(x)$ for all $x \in  \mbox{\emph{dom}}\,\sigma$. Equivalently, $\sigma \leq \tau$ if and only if $\mbox{\emph{graph}}\,\sigma \subseteq \mbox{\emph{graph}}\,\tau$ in $\mathscr{P}(X\times Y)$. \emph{(p.7, D\&P)}
\end{definition}

The first formulation says that, for $\sigma \leq \tau$, if a feature $f$ has a value in $\sigma$, then $f$ also has a value in $\tau$ and $\sigma(f) = \tau(f)$. The second, equivalent formulation says that the set of feature-value pairs defining $\sigma$ is a subset of the set of feature-value pairs defining $\tau$. So note the similarity with how the subset relation defined the lattice of set unions and intersections given earlier.

\begin{figure}
\begin{center}
     \makebox[\textwidth]{\includegraphics[width=13cm]{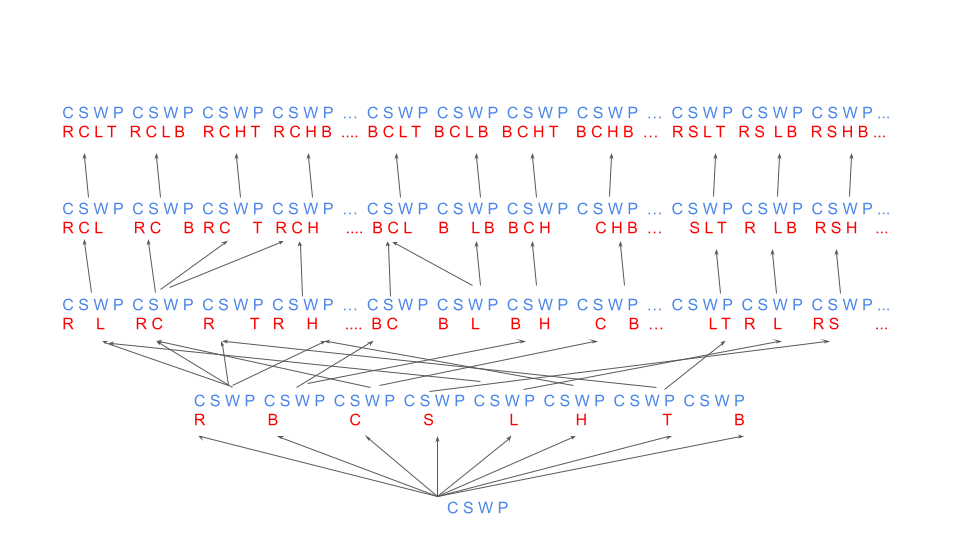}\hspace*{0cm}}
    \caption{Feature (meet-)semilattice with the subset relation over feature-value pairs (partial maps).}
    \label{fig:feature_lattice}
\end{center}
\end{figure}

Figure~\ref{fig:feature_lattice} shows part of the Hasse diagram for the set of feature values listed at the start of this subsection. It is incomplete in that not all nodes are shown, and, for those nodes that are in the diagram, not all links between nodes are shown. The blue letters correspond to the 4 features, and the red letters to the values. 
Note again the convention---following D\&P---of having the least elements towards the bottom of the page.

The lattice has a bottom element ($\bot$): the ``empty" concept at the bottom of the diagram (and if we consider each concept as a set of feature-value pairs, then it is correct to say $\bot = \emptyset$). But note how there is no single top element ($\top$), since there is no unique way of filling out all the feature values which leads to a set of feature-value pairs which is a superset of all concepts. The structure in Figure~\ref{fig:feature_lattice} is a \textbf{meet-semilattice}, since every pair of elements has a meet (greatest lower bound), but not every pair has a join (least upper bound).

It's perhaps worth giving the bottom element a name in our case, so let's call it the \emph{universal concept}. 

\begin{definition}
The \textbf{universal concept} is a partial map $\sigma$ from $\mf{Feat}$ to $\mf{Val}$ where \emph{dom}$\,\sigma = \emptyset$.  
\end{definition}

\noindent
Earlier we referred to this concept as the ``empty" concept, but ``universal" concept is better since intuitively this concept applies to all instances.

How do we combine pairs of concepts in this structure? For the equivalent of set intersection, which corresponds to going down the page, follow the links from the two concepts until they meet (which is guaranteed to happen in a meet-semilattice). Since any node can have more than one child in the diagram, there are potentially many paths to follow, and potentially many ways for paths to cross, but there will be one unique meeting point which is the highest in the lattice (since uniqueness of the greatest lower bound is guaranteed).

For the equivalent of set union, follow the links from the two concepts upwards until they join (which is not guaranteed to happen in a meet-semilattice). If there are no joining points, then the operation is undefined. If there is at least one joining point, then one of them will be uniquely the lowest in the lattice.

\subsubsection{Set Difference}

In the case where there are no clashes in two concepts (i.e. no features getting different values), then set difference is defined as one would expect (as the set difference of the sets of feature-value pairs). What about when there are clashes? For example, what is the result of $\mbox{\textsc{Cannonball}} \setminus \mbox{\textsc{Balloon}}$, where the concepts are, respectively: 
\[ \{ \langle\mf{Color,Black}\rangle, \langle \mf{Shape,Circle} \rangle, \langle \mf{Weight, Heavy}\rangle\} \;\setminus \] 
\[ \{\langle \mf{Shape,Circle} \rangle, \langle \mf{Weight, Light}\rangle\} \, ? \]

\noindent
Section~\ref{sec:set_diff} provided the solution, in terms of relative complements. First we take the meet of the two concepts, which is $\{ \langle \mf{Shape,Circle}\rangle \}$:
\[ \{ \langle\mf{Color,Black}\rangle, \langle \mf{Shape,Circle} \rangle, \langle \mf{Weight, Heavy}\rangle\} \wedge  \{\langle \mf{Shape,Circle} \rangle, \langle \mf{Weight, Light}\rangle\}\] 
$ \;\;\;= \{ \langle \mf{Shape,Circle}\rangle \} $\\

\noindent
And then find the complement of $\{ \langle\mf{ Shape,Circle}\rangle \}$ relative to \textsc{Cannonball}:
\[ \mbox{compl. of } \{ \langle\mf{ Shape,Circle}\rangle \} \mbox{ rel. to } \{ \mbox{}\langle\mf{Color,Black}\rangle, \langle \mf{Shape,Circle} \rangle, \langle \mf{Weight, Heavy}\rangle\} \]
$ \;\;\;= \{ \langle\mf{Color,Black}\rangle, \langle \mf{Weight, Heavy}\rangle \} $\\

Note that set difference is ``well-behaved" here also, as was the case for intersection (but not union): for concepts $C$ and $D$,  $C \setminus D$ is the set of feature-value pairs that are in $C$ but not $D$. 

\subsubsection{A Note on $\top$}

As discussed, the feature lattice in Figure~\ref{fig:feature_lattice} does not have a top. We could just insert one, with all the total maps from $\mf{Feat}$ to $\mf{Val}$ (the top row in the diagram) pointing at $\top$. Then we would have a lattice, and not just a semilattice.
One argument for adding a top element comes from considering the corresponding set of instances (the extension of a concept). Here it is natural to think of the top element---which we might call the impossible concept---as having the empty set as its corresponding set of instances. However, $\top$ would not correspond to a legitimate concept, since it would not be a graph of a partial map; hence we choose not to have a top in our feature lattices.

\subsubsection{A Note on Terminology}
\label{sec:terminology}

D\&P do not name the $\leq$ ordering relation in Definition~\ref{def:pmap_order}. It would be useful to have a name, since we're going to consider ways to extend this relation below. One option is to adopt the name used in other areas of computer science, including logic programming and computational linguistics, which is \emph{subsumption}, usually denoted $\sqsubseteq$.

We can also name the equivalent of the set union and intersection operations, again since we'd like to extend these to the case where the feature values themselves are ordered (and not just related through equality). The extension of the union operation is called \emph{unification}.

\begin{definition}
The unification of two feature structures (partial maps) $F$ and $F'$, $F \sqcup F'$, is the join of $F$ and $F'$ in the set of feature structures ordered by subsumption (\/$\sqsubseteq$). If $F \vee_\sqsubseteq F'$ is undefined, we say the unification has failed.
\end{definition}

We can also define an extension of the intersection operation. There does not appear to be an accepted term in the literature, but one candidate is \emph{generalisation}.

\begin{definition}
The generalisation of two feature structures (partial maps) $F$ and $F'$, $F \sqcap F'$, is the meet of $F$ and $F'$ in the set of feature structures ordered by subsumption (\/$\sqsubseteq$).
\end{definition}

\subsubsection{Some Useful Intuitions}
\label{sec:intuitions}

One reasonable question to ask at this point is: if a partial map is just a set of feature-value pairs, then why are the relevant operations not just the usual set union, intersection and difference on those sets of pairs? In fact, for the equivalent of intersection (generalisation) and set difference, they are: the generalisation of two concepts $X$ and $Y$ just is the intersection of the corresponding sets of pairs; similarly for set difference (as we have seen in the examples above). In terms of Fig.~\ref{fig:feature_lattice}, ``moving down" the diagram maintains these set operations.

The only change is when ``moving up" the diagram, when applying unification (or taking unions). Since the values for a feature can differ across concepts, simply taking the union could result in a set of feature-value pairs where one feature has more than one value, which is not a partial map over $\mf{Feat}$, and so not a concept. 

The feature structures we have been describing are an instance of a well-studied structure in pure mathematics, which turns out to have many applications, namely an \emph{intersection structure}. 
D\&P (p.48) provide a definition and more discussion of the potential applications.
As D\&P say:

\begin{quote}
    Intersection structures which occur in computer science are usually topless while those in algebra are almost invariably topped. In a complete lattice $\ldots$ of this type, the meet is just set intersection, but in general the join is not set union. (p.48, D\&P)
\end{quote}

\subsubsection{Values as Disjunctions of Feature Values}
\label{sec:disjunctions}

We finish off this section by considering disjunctive feature lattices with a finite set of values.
We might want to specify that a feature can have one value or another. For example, a swan can be black or white; a snooker ball can be white, red, yellow, green, brown, blue, pink or black. In the case of finite feature values, the natural way to represent the alternatives is with a set, in which case the value space becomes a power set.

\begin{definition}
Let $\mf{Val}$ be a finite set of values, then the disjunctive feature set associated with $\mf{Val}$, $\mf{Val}_\mf{disj} = \power{\mf{Val}} \setminus \emptyset$. $\mf{Val}_\mf{disj}$ is ordered according to the superset relation.
\end{definition}

\begin{figure}
\begin{center}
     \makebox[\textwidth]{\includegraphics[width=13cm]{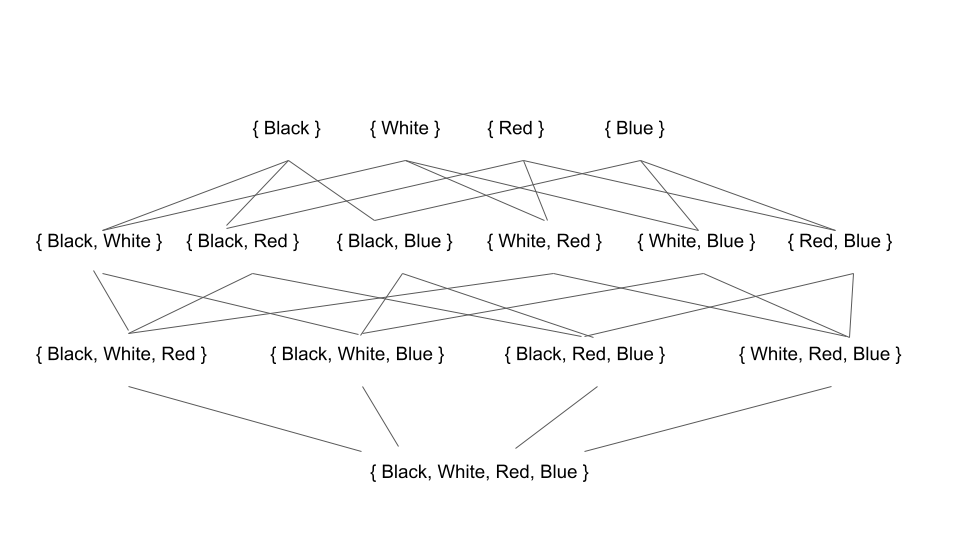}\hspace*{0cm}}
    \caption{The meet-semilattice for $\mf{Color}_\mf{disj}$, with the set of possible colors, $\mf{Color} = \{ \mf{Black, White, Red, Blue} \}$.}
    \label{fig:disj_lattice}
\end{center}
\end{figure}

Note that the disjunctive set of values does not contain the empty set, and the ordering is according to the superset, not subset, relation. Figure~\ref{fig:disj_lattice} shows the meet-semilattice defined by this ordering for an example value set of discrete colors. Why is the ordering the superset, and not subset, relation? If we consider the corresponding sets of instances (the concept extensions), then we'd like these sets to get smaller when moving ``up" the ordering. Since the sets in Figure~\ref{fig:disj_lattice} represent disjunctions, the sets of instances do get smaller when going from the bottom to the top of the semilattice.

The semilattice does have a bottom---the universal set of colors---but no top. Why not  have the empty set as the top of the semilattice, and then have the full powerset lattice? The reason is that we'd like some unifications (finding joins) to be undefined, for example:

\[ \{\langle \mf{Color},\{ \mf{Black, White} \}\rangle  \}  \sqcup \{\langle \mf{Color},\{ \mf{Red, Blue} \}\rangle\} = \mf{Undefined}.\]

\noindent
If the color semilattice had a top, the unification above would be defined, with the resulting color value being the empty set.\footnote{It's interesting to consider what the interpretation of the empty set would be in this case, if we were to add it: might it signify the ``positive" absence of color, or a concept where the lack of color is important (as opposed to color simply not being specified)?}

Consider again the above case, but this time with generalisation (finding the meet) rather than unification:

\[ \{\langle \mf{Color},\{ \mf{Black, White} \}\rangle  \} \, \sqcap \, \{\langle \mf{Color},\{ \mf{Red, Blue} \}\rangle\} =\]
\[\{\langle \mf{Color},\{\mf{Black, White, Red, Blue} \}\rangle\}.\]

\noindent
The value of the color feature is now the universe of colors, i.e. any color is possible. This case is interesting because we can ask whether a disjunction over the universe of colors is any different semantically to the unspecified feature value in our original feature lattice (Figure~\ref{fig:feature_lattice}). Or, to give a particular example, what is the difference between the following two concepts, where $\mf{Unspecified}$ means there is no feature-value pair for this particular feature:

\[ \{\langle \mf{Color},\{\mf{Black, White, Red, Blue} \}\rangle\} = \{\langle \mf{Color}, \mf{Unspecified}\rangle\} ? \]

If the answer is that there is no difference, then the feature semilattice in Figure~\ref{fig:feature_lattice} can be replaced with one where all concepts are fully specified---so there is only the top row in the diagram---but the values themselves have a semilattice structure, as in Figure~\ref{fig:disj_lattice}. (Imagine filling in all the empty slots in each partial map with the universe of colors.)
One possible reason to maintain a difference, and hence maintain both lattice structures, is to argue that semantically there is a difference between a concept where the color is important, but it can be any color, and one where the color is irrelevant and hence unspecified. A counter-argument is to note that the extensions---the set of instances to which the concepts apply---are the same in both cases. We choose to keep the partial maps since these extend naturally to the conceptual lattices in later sections, where there may not be a universal set covering all values (for example there is no finite real interval covering all possible intervals; see Section~\ref{sec:values_ranges}).

%% file: instances.tex
\section{Abstractions over Instances}
\label{sec:instances}

The standard presentation of feature lattices in the previous section assumed a finite set of features and values. But where do these features come from? Classical approaches to AI have typically assumed that the features are either provided manually by a knowledge engineer (Ch.8, \citeA{russell_norvig}) or induced automatically from text-based resources (e.g. \shortciteA{banko2007}). However, it is now recognised that one of the difficulties in the whole GOFAI enterprise was the challenge in connecting such features to perception and action \cite{cantwell-smith}, which is necessary in order that the features be properly \emph{grounded} \cite{harnad}.\footnote{The recent successes of large-scale language models such as GPT-3 suggest that such grounding may not be necessary for purely text-based applications such as machine translation or summarisation, but if the goal is an embodied, situated agent perceiving and acting in a world, then clearly its language use needs to be grounded in that world.} 

Recent progress in unsupervised representation learning, especially techniques such as the variational autoencoder (VAE) \shortcite{rezende14,kingma14}, holds promise for inducing features automatically from raw perceptual data. In this report we consider VAE and its variants, such as $\beta$-VAE \cite{beta-vae}, to provide the representation space over which abstraction takes place and upon which conceptual spaces are built. $\beta$-VAE in particular has been designed with an inductive bias towards separable representations.

The \emph{representation space}, or \emph{instance space}, is the internal space that an intelligent agent uses to represent its environment. We will assume that the representation space has separable dimensions (or more generally separable subspaces), which is necessary for the process of abstraction---dropping whole feature dimensions---to be meaningful and useful for building combinatorial abstractions. Since we are considering the representation space to be defined by a $\beta$-VAE, we assume that points in the space are arrays of real values.\footnote{In general, we may not wish to restrict the representation space in this way. For example, we may have a representation learning process which provides discrete feature values, rather than reals, but from here onwards we only consider real-valued spaces.}

\begin{definition}
\label{def:instances}
An \textbf{instance} in representation space is a point $x \in \mathbb{R}^K$.
\end{definition}

The probabilistic nature of the VAE means that the instances have some uncertainty associated with them. For now, we consider the deterministic case in which instances are known with certainty; suggestions for how to deal with a posterior distribution over instances are given in Section~\ref{sec:probs}.

Instances correspond to the situations to which concepts apply, and, as we will see later, provide \emph{extensions} to concepts. Why stop at representational instances? Do we not need extensions to be situations in the world? One reason to have instances be agent-internal is that this fits 
with our philosophy of a mentalist semantics where ``meanings are in the head". A second reason is that concepts can be generated on the fly, such as \textsc{PinkElephant}. If extensions were in the world, the extension of this concept would be the empty set.

Once we are committed to this mentalist stance, we need to explain how concepts are grounded in the world. Here we say that concepts inherit their grounding through the grounding of the internal representations. The agent's representations are grounded in the world, since they are derived through its interaction with the world. Since concepts are abstractions over representational instances, and instances are grounded through the representational mechanisms of the agent, then concepts are grounded also.

Since the concepts we are considering  are induced from the immediate perceptual experience of an agent, we call these \emph{base concepts}. The features that are likely to be induced by a representation learning algorithm such as $\beta$-VAE will be the basic features underlying color, shape, position, and so on.\footnote{The examples in this report tend to focus on the visual modality, but the representation space can cover a number of modalities, including action/motor/force representations, representation of internal body states, and representations of emotions.} \citeA{gardenfors} uses the term \emph{quality dimensions} for such features, with examples such as $\mf{temperature, brightness, weight}$, and $\mf{pitch}$. There is no suggestion that more complex concepts such as \textsc{Tiger} would necessarily be members of the conceptual lattices described in this report. Rather, complex concepts would need to be built up out of the base concepts, which we consider to have the combinatorial properties needed to create the exponentially large set of concepts that make up the human conceptual system. The meets and joins of the conceptual lattices provide a minimal mechanism for combining base concepts and moving beyond single elements from the basic feature set (consider the \textsc{Cannonball} example from earlier), but the question of what the more complex combinatorial processes are that would be required to build \textsc{Tiger} is left for future work, as is an account of how base concepts, founded on perception, can form the basis for abstract concepts such as \textsc{Democracy} \cite{lakoff_johnson}.

One last point is that the presentation here is not intended to provide a recipe for how to develop base conceptual spaces in practice; it is merely explanatory by showing what the possible spaces are in theory. It is likely that some of the spaces, e.g. the one based on arbitrary sets of points, will not be very useful to an actual agent.

\subsection{Abstraction gives Point Concepts}

\begin{figure}
\begin{center}
     \makebox[\textwidth]{\includegraphics[width=13cm]{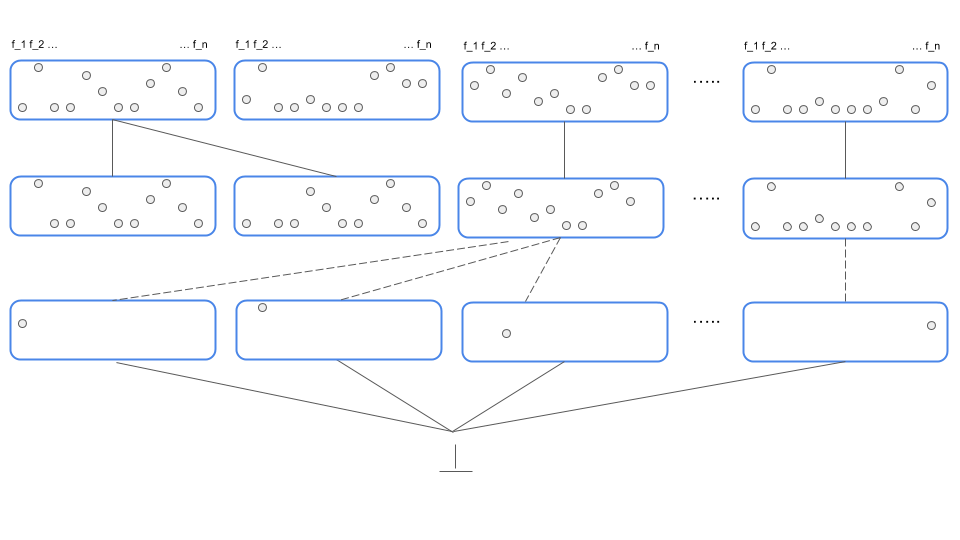}\hspace*{0cm}} \vspace*{-1cm}
    \caption{The feature semi-lattice with single points as values, and some features removed, giving point concepts. Each box represents a set of feature-value pairs, with the finite set of features across the top, and the circles representing real values on the vertical axis (assumed to extend indefinitely up and down).}
    \label{fig:real_equals_domain}
\end{center}
\end{figure}

Figure~\ref{fig:real_equals_domain} shows the meet-semilattice that results when features are removed from real-valued instances. The top row contains the instances (of which there are infinitely many), which are fully-specified feature structures where each feature dimension $f_i$ has a real value (the small circles in the diagram). The second row contains all those concepts where a single feature has been removed, and so on down to the bottom row containing concepts with just a single feature specified, and the bottom element sitting beneath all the concepts in the partial order. In fact, this picture is essentially the same as that in Figure~\ref{fig:feature_lattice}, but with the discrete feature values replaced with real numbers. We will refer to an element in this real-valued lattice as a \emph{point concept}.

What might be wrong with this picture as a proposal for a conceptual lattice that an agent could use in practice? Consider two instances $I$ and $J$ where the values on some feature dimension are close, say 3.15 and 3.16 on a ``color" dimension.\footnote{Since the feature dimensions are being discovered automatically by a representation learning algorithm, the interpretation of a dimension as corresponding to some aspect of color, say hue, would need to be carried out by a human through analysis of the representation space.} If we take the meet of $I$ and $J$, which is the operation that determines what two instances have in common (set intersection in the discrete case), then, assuming that no other feature values are equal, the result would be the empty set, or the $\bot$ concept. However, it feels as though two values which are this close (assuming a suitable range of color values) should be treated as equal as far as any concept discovery algorithm is concerned.\footnote{Finding meets is assumed in this report to be the operation that would form the basis of any concept discovery algorithm, given a set of instances as input, but in practice a number of other factors would need to be taken into account, such as how useful a potential concept is to an agent.}
The difficulty is that, whilst the VAE has provided us with feature dimensions for which abstraction is meaningful---i.e. it has provided some useful structure ``across" the feature dimensions---there is very little structure ``along" or ``within" the feature dimensions themselves, each of which is just the real number line. A natural mechanism for injecting some structure into the feature values is to group them into sets.

\subsection{Grouping Feature Values into Sets}

There are many ways in which real values can be grouped into sets. In order to maintain the lattice structure with any grouping mechanism, we need to a) define a partial order on the value sets themselves; and b) extend the definition of the partial order on partial maps (Defn.~\ref{def:pmap_order}). Let's extend the definition first:

\begin{definition}
\label{def:pmap_order_2}
The set of partial maps $(X\pmap Y)$ is ordered as follows. Given $\sigma,\tau \in (X\pmap Y)$, define $\sigma \leq \tau$ if and only if $\mbox{\emph{dom}}\,\sigma \subseteq \mbox{\emph{dom}}\,\tau$ and $\sigma(x) \leq \tau(x)$ for all $x \in  \mbox{\emph{dom}}\,\sigma$.
\end{definition}

All that has changed is that the equality relation between the values $\sigma(x), \tau(x)$ has been replaced with an ordering relation.
Note that, because of this change, the equivalent formulation in terms of subsets of the corresponding graphs (sets of feature-value pairs) no longer holds.

\subsubsection{Values as Arbitrary Sets of Reals}
\label{sec:arbitrary_sets}

Perhaps the simplest grouping mechanism is to have just arbitrary finite sets of reals as feature values.
If we were to use the terminology introduced in Section~\ref{sec:terminology}, then for b) above we would say that the subsumption relation over features needs defining also for these value sets. In the case of arbitrary sets of reals, the natural subsumption ordering is given by set inclusion. As an example for our hypothetical color dimension, the set of reals corresponding to $\textsc{DarkRed}$ would lie \emph{above} the set of reals corresponding to \textsc{Red} (assuming that the former is a proper subset of the latter).

What happens to meets and joins in this case? It is useful now to think of meets and joins happening ``within" each feature dimension, as well as ``across" dimensions. Within each feature dimension, meets are given by set union and joins by set intersection. Note that this is effectively in the ``opposite direction" to meets and joins ``across" the features (for the discrete value case), where intersections ``happened up the page" and unions ``down the page". The reason that the value sets get larger when moving down the page is that larger sets correspond to larger sets of instances, and extensions grow larger when moving down the order. The limiting case of an instance, which can be thought of as a singleton set for each dimension, is at the top of the order.

One useful intuition to take from this space is how concepts are \emph{disjunctions of point concepts}. So a point concept is a conjunction of feature values, where the feature values are real numbers, and the conjunction is being taken ``across" the feature dimensions. The disjunctions happen ``within" the dimensions, and arise from grouping values into sets. To give another color example, consider again the concept \textsc{red} corresponding to a set of reals on a single color dimension, then \textsc{red} can be thought of as a disjunction of all particular shades of red (which are all point concepts).

Another useful intuition is that the concepts are efficient descriptions of concept \emph{extensions}, i.e. the set of instances to which a concept applies; later in Section~\ref{sec:intensions_extensions} we'll call these descriptions concept \emph{intensions} (and we will also formally define a concept's extension in Section~\ref{sec:CPOs}). The descriptions are efficient because, for the missing dimensions in any abstraction, there is an assumption that all points along those dimensions form part of the extension, so there is no need to explicitly store or represent those points.

Let's consider taking the meet of two instances in this conceptual space. Let the two instances be
\[\mbox{Inst1} = \{ (f1,u1), (f2,u2),\ldots, (f\mbox{n},u\mbox{n}) \}, \mbox{Inst2} = \{ (f1,v1), (f2,v2),\ldots,(f\mbox{n},v\mbox{n}) \},\] then \[\mbox{meet(Inst1,Inst2)} = \{ (f1,\{u1,v1\}), (f2,\{u2,v2\}),\ldots,(f\mbox{n},\{u\mbox{n},v\mbox{n}\}) \}.\]

First note how the instances are ``complete" point concepts (without any abstraction having taken place), and how the resulting concept from the meet operation can be represented as a 
disjunction of point concepts (in this case a disjunction of instances).\footnote{Note that the instances in the disjunction are not just Inst1 and Inst2, but all instances formed by taking conjunctions of the $u$'s and $v$'s.} An equivalent way to represent the concept is not as a disjunction of conjunctions, but as a conjunction of disjunctions, which shows how the values along each feature dimension are grouped into sets.
The meet can be written as \[(u1 \textsc{ or } v1) \textsc{ and } (u2 \textsc{ or } v2) \ldots \textsc{ and } (u\mbox{n} \textsc{ or } v\mbox{n}).\]

\begin{figure}
\begin{center}
     \makebox[\textwidth]{\includegraphics[width=13cm]{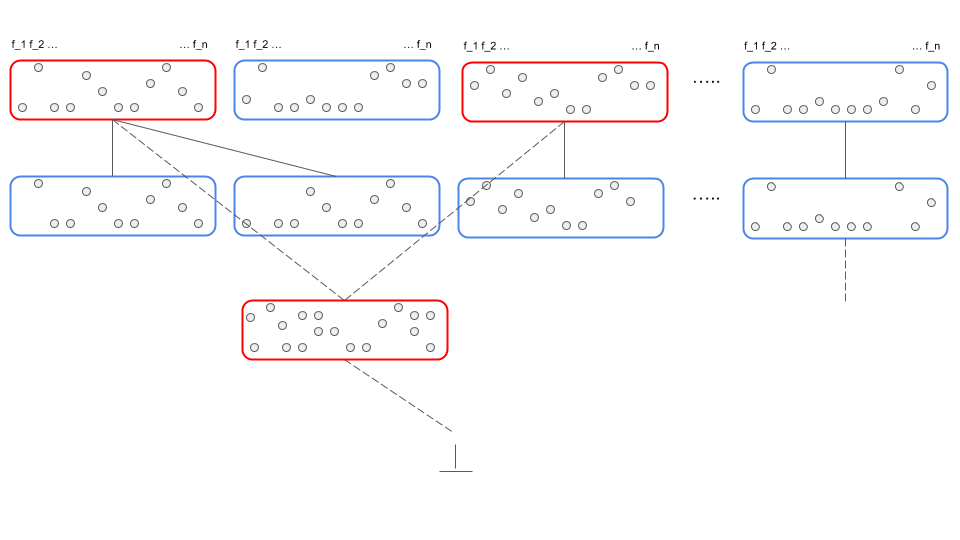}\hspace*{0cm}}
    \caption{The meet of the two instances in the red boxes (given by set union along the dimensions) when there are no restrictions on the value sets; no abstraction takes place.}
    \vspace*{-0.2cm}
    \label{fig:meet_no_abstraction}
    \end{center}
\end{figure}

\begin{figure}
\begin{center}
     \makebox[\textwidth]{\includegraphics[width=13cm]{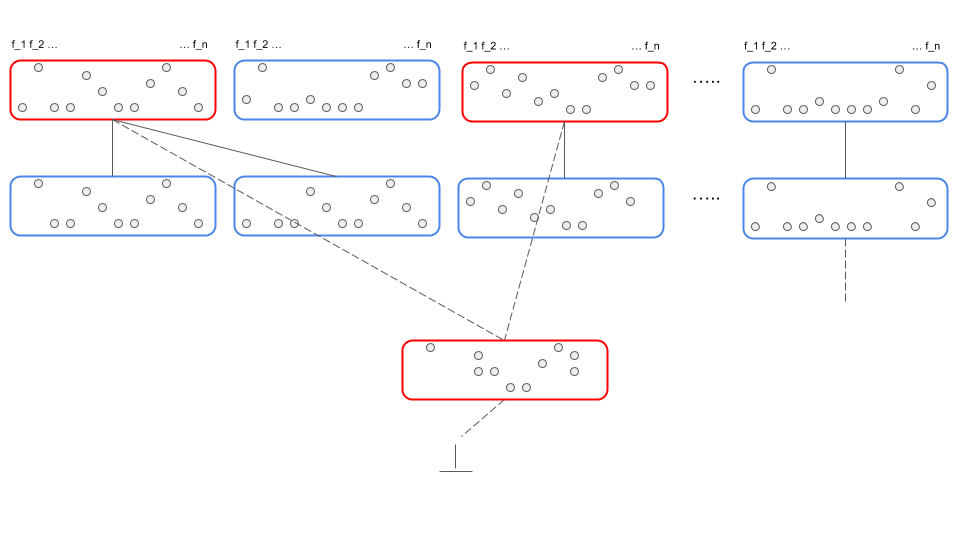}\hspace*{0cm}}
    \caption{The meet of the two instances in the red boxes when the value sets only contain elements that are sufficiently close; abstraction now takes place since whole feature dimensions are dropped.}
    \vspace*{-0.2cm}
    \label{fig:meet_w_abstraction}
    \end{center}
\end{figure}

The meet operation has grouped the values into sets, but note that no abstraction has taken place, since none of the dimensions have been dropped. The problem, from the perspective of wanting to find useful abstractions, is that no restrictions have been placed on the value sets.
A natural restriction is to only allow sets as values in which the maximum and minimum elements are within some distance $\epsilon$ of each other. That way, only values which are relatively close, or ``similar", will get grouped together. If we now assume, for this particular case, that $\vert u\mbox{i} - v\mbox{i}\vert < \epsilon$ for $i = 1$ only, then $\mbox{meet(Inst1,Inst2)} = \{ (f1,\{u1,v1\}) \}$. Notice how placing this restriction on the feature values has neatly led to a process of abstraction happening ``for free" when performing the meet operation. Figure~\ref{fig:meet_no_abstraction} shows the outcome of the meet operation when there is no restriction on the sets, whereas Figure~\ref{fig:meet_w_abstraction} shows what happens when the sets only contain values that are sufficiently close.

Are there any limitations of this conceptual space? Meets appear to work well (with the maximum distance requirement), and the grouping operation adds minimal additional structure to the original instance space. There is a problem, however, with the join operation. To provide some intuition, and adapting an example from earlier, suppose that we have the following two concepts:
\[ \{ \langle \mf{Color,Black_1}\rangle, \langle \mf{Shape,Circle} \rangle \}, \,\,\{\langle \mf{Color,Black_2}\rangle, \langle\mf{Weight, Heavy} \rangle \}. \] 
We'd like to unify these two concepts, i.e. take the join, in order to create the concept \textsc{Cannonball} (a heavy black circle). However, the join will only be defined if $\mf{Black_1} \cap \mf{Black_2}$ is non-empty. Note that meets work fine, even in the general case: the meet of these two concepts is $\{ \langle \mf{Color,Black_1 \cup \mf{Black_2}}\rangle\}$. But for the joins, it feels as though an additional assumption is needed to get the outcome we'd like. If $\mf{Black_1}$ and $\mf{Black_2}$ are disjoint, but contain values that are close, then we'd like the unification of these two concepts to be well-defined.

\subsubsection{Values as Closed Intervals of Reals}
\label{sec:values_ranges}

We made the assumption above that only points that are sufficiently close along a dimension can form part of a concept. A natural extension of that assumption is the following:

\begin{proposition}
{\bf Convexity condition:} If points $A$ and $B$ (within a feature dimension) both form part of a concept $C$, then all points between $A$ and $B$ are also part of $C$.
\label{prop:convex}
\end{proposition}

There is a close link here with \citeA{gardenfors}, who also imposes a convexity condition (see Section~\ref{sec:gardenfors} below). However,  G{\"a}rdenfors assumes such a condition applies to all concepts, whereas we are only applying it to \emph{base} concepts (as explained in the introduction to this section). There is no suggestion that more complex concepts need to be convex; in fact, the disjunctive and negated concepts described below, for example, are not.

Proposition~\ref{prop:convex} leads to feature values which are closed intervals of real values, i.e. $\mf{Val} = \{ [x,y] \,\vert\, x,y \in \mathbb{R}, x \leq y$\}, with the following ordering.

\begin{definition}
\label{def:range_order}
The set of real-valued intervals is ordered as follows. Given $\rho,\pi \in \{ [x,y] \,\vert\, x,y \in \mathbb{R}, x \leq y \}$, define $\rho \leq \pi$ if and only if $\rho \supseteq \pi$. Let $\rho = [\rho_1,\rho_2]$ and $\pi = [\pi_1,\pi_2]$, then equivalently $\rho \leq \pi$ if and only if $\rho_1 \leq \pi_1$ and $\pi_2 \leq \rho_2$. 
\end{definition}

\noindent
So if we think of an interval as a set of real numbers, then one interval $\rho$ is below another $\pi$ in the order if and only if $\pi$ is a proper subset of $\rho$, which means that $\pi$ is fully contained within $\rho$.

One potential confusion here is that, if we think of the ordering on the features in Definition~\ref{def:pmap_order_2}, i.e. the ordering on the domains of the maps, then the domains become more specific as the domains get larger; whereas, for the values as intervals, the ordering is the reverse: intervals which are larger and contain other intervals are less specific, and hence earlier in the ordering (as was the case for the arbitrary sets of points). One way to clarify the confusion is to consider how informative the respective sets are. On the features side, a concept which has many features is more informative than one that doesn't, in the sense that it corresponds to a smaller set of possible instances, and determines more aspects of the instance space.
Conversely, a feature value with a large range is less informative than a value with a small range, since the larger range corresponds to a larger set of possible instances.

What happens to meets and joins in this case?
The meet of two intervals (within a feature dimension) is the convex hull and the join is the intersection (leading to an undefined join if the intersection is empty). 

\begin{proposition} \label{prop:interval_unification}
Let $\mathcal{R}$ be the set of real-valued intervals, $\mf{Feat}$ a finite set of features, and $\mf{Val} = \mathcal{R}$. Assuming the ordering on $\mathcal{R}$ from Definition~\ref{def:range_order}, and given $F = \{ (f_i,v_i) \,\vert\, f_i \in \mf{Feat}, v_i \in \mf{Val}\}_i$ and $F' = \{ (f'_i,v'_i) \,\vert\, f'_i \in \mf{Feat}, v'_i \in \mf{Val}\}_i$, then $F \sqcup F'$ (if defined) is:
\[ \{ (f_i,v_i)\,\vert\,f_i \in \mbox{\emph{dom}}\,F, f_i\notin\mbox{\emph{dom}}\,F' \} \,\cup\, \{ (f'_i,v'_i)\,\vert\,f'_i \in \mbox{\emph{dom}}\,F', f'_i\notin\mbox{\emph{dom}}\,F \} \,\cup\,\]
\[\{ (f_i,v_i \cap v'_i)\,\vert\,(f_i,v_i) \in F, (f_i,v'_i)\in F' \} \]
\end{proposition}

The unification $F \sqcup F'$ will be undefined if $F$ and $F'$ share a feature where the corresponding values do not overlap (i.e. the intersection of the corresponding intervals is the empty set). Where $F$ and $F'$ share a feature where the corresponding values do overlap, then the value of the unification for that feature will be the intersection of the two intervals.

\begin{proposition} \label{prop:interval_generalisation}
Assuming the ordering on $\mathcal{R}$ from Definition~\ref{def:range_order}, and $F$ and $F'$ as above, then $F \sqcap F'$ is:
\[\{ (f,\left[\min\{a_1, a_2\}, \max\{b_1, b_2\} \right])\,\vert\,(f,[a_1,b_1]) \in F, (f,[a_2, b_2])\in F' \}\]
\end{proposition}

The generalisation $F \sqcap F'$ will be the empty set, i.e. the universal concept, if $F$ and $F'$ do not share any features in common. Where $F$ and $F'$ do share a feature, then the value of the generalisation for that feature will be the convex hull of the two intervals, i.e. the smallest interval containing both.

In order for the meet operation to have the capacity to return an abstraction, i.e. for some dimensions to be dropped, the intervals again need to be restricted in length, as for the arbitrary sets of points. In this case, the meet only returns an interval if the convex hull is within the limit (and if no convex hull is within the limit across all dimensions then the meet is the universal concept at the bottom of the ordering).

Section~\ref{sec:disjunctions} described a conceptual space of disjunctive concepts when the feature values are discrete. In this section we extend that analysis to the case of closed-interval values, as well as negation.

\paragraph{Disjunctive Concepts}

The first point to make is that all the conceptual spaces we have considered so far are essentially disjunctive, in the sense that all concepts can be thought of as ``conjunctions of disjunctions", as well as disjunctions of point concepts, as described in Section~\ref{sec:arbitrary_sets}. The fact that the conceptual space has already been factorised into separable dimensions immediately leads to this form. 

So how do we add disjunctive concepts to the space of interval values? Since we can have disjunctions of point concepts, then the space of arbitrary sets of points is required. Note how this contains all sets of closed intervals, allowing disjunctions of property intervals, for example \textsc{green or blue or red} (assuming each color is a closed interval on some color dimension). However, didn't we describe a problem when taking joins in such a space, leading to the convexity requirement? If so, we wouldn't necessarily want the disjunctive space to be an agent's primary conceptual space for concept discovery and reasoning.

The answer is to only take meets, as part of some concept discovery process, on the original space containing the closed intervals. Intuitively this makes sense: suppose I have two similar instances $I$ and $J$ (along some dimension, i.e. $I,J \in \mathbb{R}$). If I'm allowed to form disjunctive concepts during the initial discovery phase, then $\{ I, J \}$ would be a reasonable concept to form. However, if the convexity condition is being applied, then $[I, J]$ would be the outcome. Suppose further that there are two similar instances $K$ and $L$, which are distant from $I$ and $J$, then again we can form $[K, L]$, but it is only \emph{after} forming these two concepts that we can consider forming the disjunctive concept $\{ [I, J], [K, L] \}$.

To provide some intuition about how this space works, consider the following set of concepts, all represented as closed intervals along a single color dimension:

\[ \{ \textsc{blue}, \textsc{red}, \textsc{green}, \textsc{dark-red} \} \]

\noindent
Suppose further that all these concepts are disjoint (e.g. $\textsc{blue} \cap \textsc{red} = \emptyset$), except \textsc{red} and \textsc{dark-red} where $\textsc{dark-red}  \subset \textsc{red}$. Here are some examples of meets ($\wedge$) and joins ($\vee$) in this space:

\[ \textsc{blue} \vee \textsc{red} = \textsc{undefined}\]
\[ \textsc{blue} \wedge \textsc{red} = \{ \textsc{blue}, \textsc{red} \} \]
\[ \textsc{red} \vee \textsc{dark-red} = \textsc{dark-red} \]
\[ \textsc{red} \wedge \textsc{dark-red} = \textsc{red} \]
\[ \{ \textsc{red}, \textsc{blue} \} \vee \textsc{dark-red} = \textsc{dark-red} \]
\[ \{ \textsc{red}, \textsc{blue} \} \wedge \textsc{dark-red} = \{ \textsc{red}, \textsc{blue} \} \]
\[ \{ \textsc{red}, \textsc{blue} \} \wedge \textsc{green} = \{ \textsc{red}, \textsc{blue}, \textsc{green} \} \]
\noindent
Note that the maximum-length condition is not being applied in any of these cases, since it does not apply to the disjunctive space.

In Section~\ref{sec:intensions_extensions} below we define a concept as a conjunction of \emph{properties}, where, for the conceptual space based on intervals, a property is a closed interval along some feature dimension. This definition can naturally be extended to the disjunctive space so that disjunctive concepts are conjunctions of \emph{sets} of properties. 

\paragraph{Negated Concepts}

Negations of concepts can be formed in the obvious way through taking set complements of specified feature values. For example, consider a concept $C$ which has the value $[c_1, c_2]$ along some feature dimension; the concept $\neg C$ will have the value $\langle (- \inf, c_1), (c_2, + \inf)\rangle$. We can also introduce this additional ``negated" space into the partial order, simply through the subset relation. Examples of relationships in this partial order ($\subseteq$), assuming the color intervals used above, include: $\textsc{red} \subseteq \neg \textsc{blue}$, $\neg \textsc{red} \subseteq \neg \textsc{dark-red}$, $\neg (\textsc{red or blue}) \subseteq \neg \textsc{red}$.

However, again we need to be careful when taking meets in this new space: consider two concepts $I$ and $J$ that are far apart; in the original space of maximum-length intervals the convex hull would be undefined, and hence the meet would be the universal concept. But with the addition of the negated space below the original space, there are now many lower bounds for $I$ and $J$. For example, there are many ways for  \textsc{red} and \textsc{blue} to now meet in the negated space, such as \textsc{not-green}, \textsc{not-yellow}, and so on, but none of these is a greatest lower bound. Again the solution is to not allow the initial concept discovery process to operate in this negated space, which again makes intuitive sense: it is only \emph{after} we have found some ``positive" concepts that we can start to consider negative ones.

\subsubsection{Values as Equivalence Classes of Points}
\label{sec:equiv_class}

As a special case of the closed-intervals conceptual space, we imposed a length restriction on the intervals so that finding meets could lead to some abstraction taking place. A further restriction would be to only allow intervals from a partition of the real number line, as described in Section~\ref{sec:discrete_feats}.\footnote{Technically the intervals are \emph{half-closed} in this case.} The partitions could be different for each feature dimension, but the key point is that there are now only a finite number of values for each feature. In fact, this corresponds to the discrete feature values case set out in Section~\ref{sec:formalisation}.

\begin{figure}
\begin{center}
     \makebox[\textwidth]{\includegraphics[width=13cm]{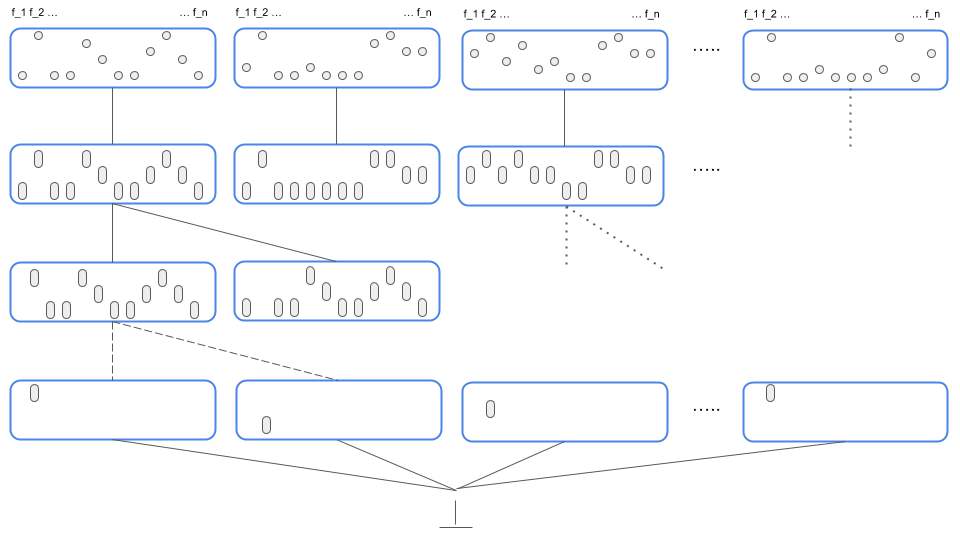}\hspace*{0cm}}
    \caption{The conceptual space with equivalence classes as values, and some features removed.}
    \label{fig:partition_domain}
\end{center}
\end{figure}

Figure~\ref{fig:partition_domain} shows the conceptual lattice in this case. Here the order at the very top of the diagram---relating instances to concepts with all features defined---is determined by the partition function, and the remaining order is given by the subset relation over feature-value pairs.
Note that there is no ordering among the values themselves, since the partition along any feature dimension results in an \emph{anti-chain} in which none of the values are related to any other values.

\subsection{Properties, Intensions and Extensions}
\label{sec:intensions_extensions}

We have defined a concept's extension as the set of instances to which the concept applies (this will be made more precise in Section~\ref{sec:CPOs}), but what about its \emph{intension}? As discussed in Section~\ref{sec:arbitrary_sets}, the efficient descriptions of a concept's extension---obtained via abstracting away whole separable dimensions---are intensional descriptions, so can we provide a definition? First, let's give a name to the values of a conceptual space, for example the arbitrary sets of points in Section~\ref{sec:arbitrary_sets} or closed intervals in Section~\ref{sec:values_ranges}. And as a reminder, here is how a concept was defined in Section~\ref{sec:formalisation}:

\begin{definition}
\label{def:concept_repeat}
(Repeat of Defn.~\ref{def:concept}) Assuming a set of features $\mf{Feat}$ and a set of values $\mf{Val}$, a concept $C$ is a map $\delta_C : \mf{SubFeat} \rightarrow \mf{Val}$, where $\mf{SubFeat} \subseteq \mf{Feat}.$
\end{definition}

The set of values is determined by the particular conceptual space in question. Examples of $\mf{Val}$ we have considered so far include equivalence classes from a partition of the real number line, arbitrary sets of real values, and closed intervals of reals (with a limited maximum length). Intuitively, these values act like \emph{properties}; to give a concrete example, assuming color can be represented on a single real-valued dimension, then $\textsc{dark-red}$ would correspond to a particular closed interval. Hence let's call these values properties:

\begin{definition}
A conceptual space is made up of a set of feature dimensions $\{\mf{Feat}_i\}_i$ and each dimension has a corresponding set of values $\mf{Val}_i$. A value $P \in \mf{Val}_i$ is a \textbf{property}.
\end{definition}

Now we can provide an alternative, but equivalent, definition of a concept (or a concept's intension) to the one given above.

\begin{definition}
\label{def:concept2}
Assuming a set of feature dimensions $\{\mf{Feat}_i\}_i$ and corresponding sets of values $\{\mf{Val}_i\}_i$, a concept $C$ is a set of properties $\mathcal{P}(C) = \{ (P,j) \vert P\in \mf{Val}_j, j\in\mathcal{F}(C)\}$, where $\mathcal{F}(C)$ is the set of features that are ``on" for concept $C$.
\end{definition}

A concept can now also be thought of as a \emph{conjunction of properties}. As a special case, a point concept is a conjunction of what we might call \emph{point properties}.

\subsubsection{Link to Formal Concept Analysis}

\begin{figure}
\begin{center}
     \makebox[\textwidth]{\includegraphics[width=6cm]{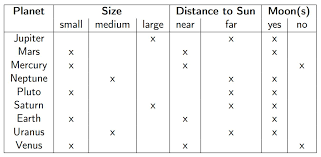}\hspace*{0cm}}
    \caption{A set of attributes for the planets (example from p.65, D\&P).}
    \vspace*{-0cm}
    \label{fig:fca}
    \end{center}
\end{figure}

In Formal Concept Analysis (FCA), a concept is defined as a pair, consisting of an intension and an extension (Ch.3, D\&P). However, in FCA the intensions---or what D\&P call \emph{intents}---are sets of binary \emph{attributes}. Figure~\ref{fig:fca} gives an example from D\&P. The extension---or what D\&P call the \emph{extent}---of an attribute is the set of objects that possess that attribute. For example, the extent of \textsc{has-no-moon} is \{\textsc{Mercury, Venus}\}.

From this simple mathematical structure, a rich theory of formal concepts emerges, built around the mathematics of partial orders from Section~\ref{sec:formalisation}. In particular, there are some intimate relationships between the topped $\cap$-structures that we mentioned briefly in Section~\ref{sec:intuitions}, closure operators (which have close connections with the CPOs described in Section~\ref{sec:CPOs}), and so-called Galois connections (p.68, D\&P). Developing these relationships for our own concepts theory would be an interesting avenue for further mathematical work.

One key difference in our treatment of concepts is that we don't assume binary (or even discrete) attributes, and do not assume that these are given in advance. The only constraints on properties that we have imposed in
this section, and that will be made more precise in
Section~\ref{sec:CPOs}, are that a) the properties on each feature dimension are ordered; and b) instances (along a feature dimension) are limiting cases of properties. 

Another difference is the relationship between intensions and extensions. In our formalisation, all combinations of available feature values define an extension; for example, $\{\langle \mf{Color,Pink}\rangle, \langle \mf{Shape,Elephant} \rangle\}$ would be a perfectly reasonable concept for us, and would have an extension defined by these two feature values (with the other features varying freely). In contrast, in FCA, the extension corresponding to this set of attributes may well be empty, depending on the objects available for making up the extensions (i.e. there may not be any pink elephants in the set of objects).

\subsubsection{Link to G{\"a}rdenfors' Conceptual Spaces}
\label{sec:gardenfors}

G{\"a}rdenfors' theory of conceptual spaces \shortcite{gardenfors} shares some similarities, and has some differences, with the theory being described here, so it is worth considering what those similarities and differences are. First, G{\"a}rdenfors emphasises the fact that his theory is a geometric theory, with any conceptual space being defined by a set of dimensions. G{\"a}rdenfors calls these dimensions \emph{quality dimensions}, but they are essentially the same---mathematically at least---as the feature dimensions we have been using. Examples of quality dimensions that G{\"a}rdenfors gives include $\mf{temperature}$, $\mf{brightness}$, $\mf{weight}$, and $\mf{pitch}$.

A key notion for G{\"a}rdenfors is the idea that quality dimensions can be either \emph{integral} or \emph{separable}. A set of dimensions is integral if ``one cannot assign an object a value on one dimension without giving it a value on another" (p.24, \citeA{gardenfors}). Examples of integral dimensions are $\{\mf{hue}, \mf{brightness}\}$ and $\{\mf{pitch}, \mf{loudness}\}$. Dimensions which are not integral are separable.
In the exposition of our concepts theory we have been assuming the notion of separable dimensions, but note that G{\"a}rdenfors' definition is different to that given in, for example, \shortciteA{higgins:disentangled}, which is based on invariant transformations of world state.

Another key notion for G{\"a}rdenfors is the idea of a \emph{natural property}, which is defined as a convex region of a domain; then a \emph{natural concept} is defined as a set of natural properties. We have also defined a base concept above as a set of properties, and the particular case of the closed-interval properties arose from a convexity condition being applied. However, there is no suggestion in our presentation that \emph{all} concepts are convex; indeed, the disjunctive and negated concepts from Section~\ref{sec:values_ranges} are examples that are not.
Another area in which our presentation differs from G{\"a}rdenfors is in our clear separation between a concept's intension and extension. And finally, G{\"a}rdenfors does not consider how the properties in his conceptual spaces could be ordered, whereas partial orders lie at the heart of our formalisation. 

\subsection{Conceptual Spaces as CPOs}
\label{sec:CPOs}

The final part of this section brings all the  lattice-theoretic ideas together into a more precise definition of a conceptual space,  by offering the following proposition, which is the main mathematical proposal of this work.

\begin{proposition}
A conceptual space of base concepts is a complete partial order (CPO) where the maximal elements of the CPO are instances in representation space, and the bottom element of the CPO is the universal concept.
\label{def:concept_space}
\end{proposition}

This subsection first provides the mathematical definitions required to understand the notion of a CPO, and then informally shows how the conceptual spaces we have defined are all examples of CPOs. The following description and definitions are taken from \citeA{abramsky_domains} (A\&J) and D\&P.

The definition of a CPO relies on the notion of a \emph{directed set}.

\begin{definition}
Let $S$ be a non-empty subset of an ordered set $P$. Then $S$ is said to be \textbf{directed} if, for every pair of elements $x,y \in S$, there exists $z \in S$ such that $z \in \{x,y\}^u$
\end{definition}
\noindent where $\{x,y\}^u$ is the set of upper bounds of $\{x,y\}$ (p.148, D\&P). Simple examples of directed sets are chains, since every pair of elements in a chain is related, and so the maximum of any two elements is an upper bound on the pair. Another example is the set of finite subsets of an arbitrary set (ordered by the subset relation), where an upper bound of any pair of subsets is provided by the union.

CPOs are partial orders in which each directed subset has a join.

\begin{definition}
We say that an ordered set $P$ is a \textbf{CPO} if
\begin{enumerate}[label=(\roman*)]
    \item $P$ has a bottom element, $\bot$,
    \item $\bigsqcup D$ $(:= \bigvee D)$  exists for each directed subset $D$ of $P$. \emph{(p.175, D\&P)}
\end{enumerate}
\end{definition}

\noindent
If (ii) is satisfied but not (i), i.e. there is no bottom element, then the partial order is often called a \emph{directed-complete partial order (DCPO)}.

A\&J (p.15) provide some examples of (D)CPOs, noting that every finite partially ordered set is a DCPO. An instructive example of a partial order that is not a DCPO is the set of natural numbers with the usual order. This set is directed, since it is a chain, and every \emph{finite} directed subset has a join (the maximum element), but the whole set itself, for example, does not have a join.\footnote{Note that the join does not have to be in the directed subset, it just needs to exist in $P$.}

There is an alternative formulation of CPOs in terms of chains.

\begin{definition}
Let $P$ be an ordered set. Then $P$ is a CPO if and only if each chain has a least upper bound in $P$. \emph{(p.176, D\&P)}
\end{definition}

This formulation will provide some useful intuition in the context of our conceptual CPOs. It also explains why every CPO has a bottom element: since each chain in a CPO has a join, then there must be a join for the empty chain (which is the empty set), and hence there must be a bottom element, since the join of the empty set is $\bot$ (p.179, D\&P). We'd like to commit to the existence of $\bot$ in our conceptual spaces, since this corresponds to what we are calling the universal concept---the concept that applies to all instances---and every conceptual space should have one.

A CPO $P$ has at least one \emph{maximal element}, i.e. Max$\,P \neq \emptyset$ (p.229, D\&P; p.73, \citeA{priestley:02}). 

\begin{definition}
Let $P$ be an ordered set, and let $Q \subseteq P$. Then $a \in Q$ is a \textbf{maximal element of} $Q$ if $a \leq x$ and $x \in Q$ imply $a = x$. We denote the set of maximal elements of $Q$ by \emph{Max}$\,Q$. \emph{(p.16, D\&P)}
\end{definition}

In our case, if the conceptual space is $P$, then Max$\,P$ is the set of instances (Prop.~\ref{def:concept_space}). What does it mean for these elements to be maximal? Intuitively, these are the elements that sit at the top of the feature (semi-)lattice, with any other element in the lattice an abstraction and/or grouping of these instances, as determined by the partial order. Hence instances are fully-specified feature value pairs, in the sense that every feature has a value, and every value is maximally informative.

As promised, we can now formally define a concept's extension.

\begin{definition}
The \textbf{extension} of a concept $C$ in a conceptual CPO $P$ is the set of instances $\{ x \in \mbox{\emph{Max}}P \,\vert\, x \geq C \}$.
\end{definition}

Section~\ref{sec:example_domains} presents our example conceptual spaces in terms of CPOs, informally demonstrating in each case how the partial order results in a CPO with the instances as maximal elements. Some of this presentation is repeated from earlier.

\subsubsection{Examples of Conceptual Spaces as CPOs}
\label{sec:example_domains}

We will start with the set of instances, and consider minimal ways in which we can form CPOs by a) abstracting away from those instances by removing features, and b) grouping the real values into sets of various kinds.

\begin{figure}
\begin{center}
     \makebox[\textwidth]{\includegraphics[width=13cm]{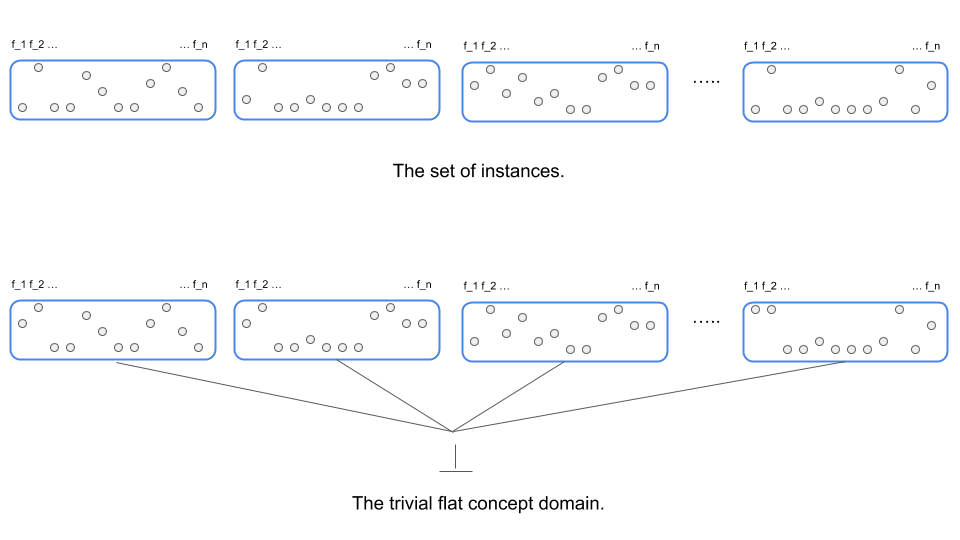}\hspace*{0cm}}
    \caption{The trivial conceptual CPO containing only the universal concept.}
    \vspace*{-0.2cm}
    \label{fig:trivial_domains}
\end{center}
\end{figure}

\paragraph{The Universal Concept}
Note that the set of instances themselves does not form a CPO, since we must have a bottom element. Hence perhaps the most trivial conceptual CPO is the one containing only the instances and the universal concept, shown in Figure~\ref{fig:trivial_domains}, which involves abstracting completely along the feature axis, by removing all features. This results in what is known as a \emph{flat} ordered set, in which each instance is only related to the bottom element (the universal concept). Note that the instances themselves are not ordered relative to each other, and hence form an \emph{anti-chain}.\footnote{Since the instances are always the maximal elements (according to Prop.~\ref{def:concept_space}), they will always form an anti-chain in any conceptual CPO.}

\begin{figure}
\begin{center}
     \makebox[\textwidth]{\includegraphics[width=13cm]{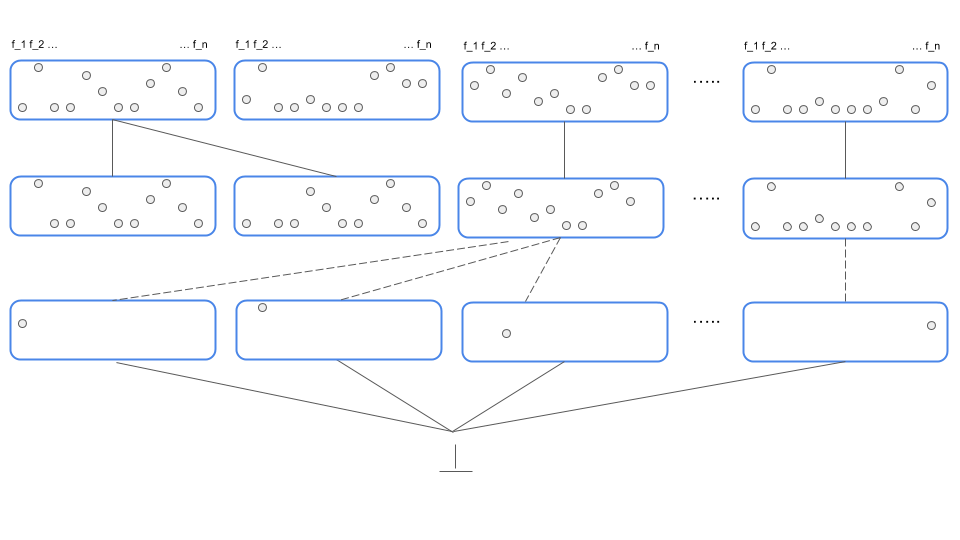}\hspace*{0cm}} \vspace*{-1cm}
    \caption{The conceptual CPO with single points as feature values, and some features removed, giving point concepts.}
    \label{fig:real_equals_domain_repeat}
\end{center}
\end{figure}

\paragraph{The Conceptual CPO of Point Concepts}
Now let's perform an abstraction as above, but this time removing just some of the features. Concepts are now single points in a subspace of the representation space, and removing features amounts to taking a projection in the representation space. The partial order is given by the subset relation over the feature-value pairs. Hence different concepts with the same set of features will form an anti-chain (since in this conceptual CPO there is no way of comparing values). Figure~\ref{fig:real_equals_domain_repeat}
(a repeat of Fig.~\ref{fig:real_equals_domain}) shows this CPO. Note how any chain---a set of concepts which are all related---has a join (the top of the chain), which is the defining characteristic of a CPO, and how each concept is below at least one instance, which are the elements at the top of the diagram.

\paragraph{The Conceptual CPO of Equivalence Classes}

\begin{figure}
\begin{center}
     \makebox[\textwidth]{\includegraphics[width=13cm]{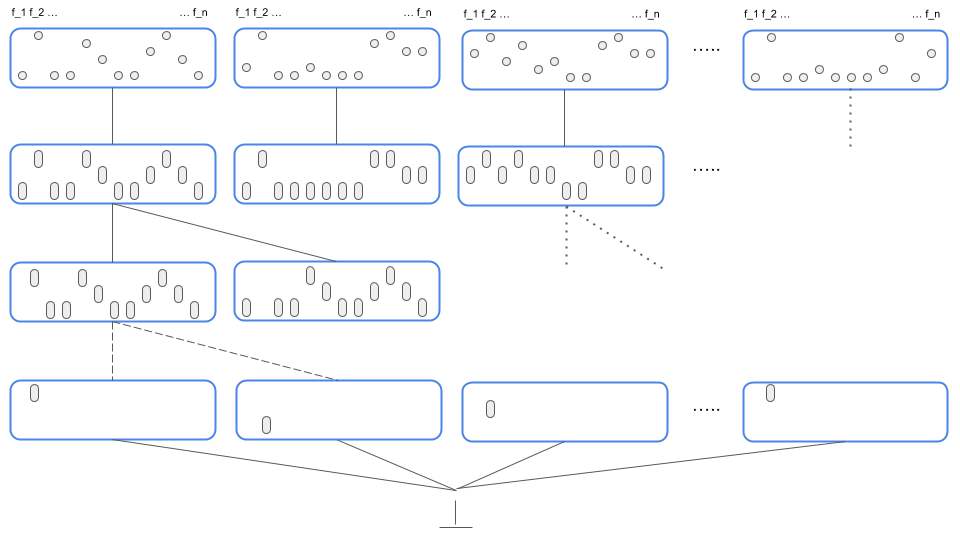}\hspace*{0cm}}
    \caption{The conceptual CPO with equivalence classes as values, and some features removed.}
    \label{fig:partition_domain_repeat}
\end{center}
\end{figure}

Now let's group values together, by putting the values on each feature dimension into equivalence classes, as described in Section~\ref{sec:equiv_class}. In fact, if we consider each feature dimension, and each bucket, separately, then this creates a flat order, like that in Figure~\ref{fig:trivial_domains}, but with the bottom element replaced with the bucket.
If we also perform abstraction by removing some features, then we have the conceptual CPO in Figure~\ref{fig:partition_domain_repeat} (a repeat of Fig.~\ref{fig:partition_domain}). Here the order at the very top of the diagram---relating instances to concepts with all features defined---is determined by the partition function, and the remaining order is given by the subset relation over feature-value pairs.
Note again how any chain in the partial order has a join (the top element of the chain), and how every concept sits below a number of instances.

This conceptual space is a useful one to consider when providing more intuition around the idea of a CPO. What are the directed subsets in this CPO? These are the sets of concepts which are all consistent with each other, i.e. there are no clashes on any of the feature dimensions. Moreover, the subset must be closed when considering upper bounds, in the sense that each pair of concepts in the subset must have an upper bound in the subset (i.e. when following links upwards in the feature lattice, there must be at least one point in the subset where the links cross). And finally, the subset itself must have a join, i.e. a least upper bound in the CPO. The join does not have to be in the subset, but in this case it always will be---it will be the maximal element of the subset---because of the ``finite" nature of the CPO.

\paragraph{The Conceptual CPO of Closed Real-Valued Intervals}

Finally we can group real values into closed intervals, and order those intervals by the inclusion (superset) relation. Note how the point values in the instances are limiting cases of the closed intervals, with a point $x \in \mathbb{R}$ corresponding to the interval $[x,x]$. This conceptual space is a special case of the more general space in which real values are grouped into arbitrary sets, where again the point values in the instances are limiting cases, with a point $x \in \mathbb{R}$ corresponding to the singleton set $\{x\}$.

An interesting extension of this theoretical section would be to demonstrate that all our conceptual CPOs are also examples of \emph{domains}, a key notion in theoretical computer science \cite{abramsky_domains}. For example, the trivial CPO in Figure~\ref{fig:trivial_domains} is an example of an \emph{algebraic domain}. In fact, all but the conceptual CPO of closed intervals are algebraic domains, with the closed-interval case a \emph{continuous domain}.

%% file: probabilities.tex
\section{Probabilities}
\label{sec:probs}

There are a number of ways in which probabilities can feature in the formalisation, all depending on how the probabilities are interpreted. First there is the posterior distribution from, for example, a $\beta$-VAE. Since the instance space is assumed to come from a representation learning method such as $\beta$-VAE, we need an account of how those distributions propagate through the conceptual spaces. Second, there can be statistical correlations across feature dimensions; to give a concrete example, the color of an apple may not be independent of its taste \cite{gardenfors}. And finally, there is a potential use for distributions in injecting some ``fuzziness" into the conceptual spaces described so far, which have all assumed hard boundaries, or binary membership conditions, for a concept's properties.\footnote{An interesting, and largely uninvestigated, question arises in the last case, which is how to order distributions \cite{wetering_msc_thesis} so that they can form a lattice structure.}
In this document we will focus on the first case, leaving an account of the potential use of probabilities for fuzziness for future work.

\subsection{Posterior Distributions from  Representation Learning}

Given an input $x$, for example an image or a video, a $\beta$-VAE defines a posterior distribution $p(z\vert x)$ over instances $z$, where the distribution is constrained to be a multivariate Gaussian with a diagonal covariance matrix. How should we  interpret this posterior? It's the uncertainty associated with the instance space, given this particular input. Note that we assume there is one true underlying $z$ which generated $x$, but the stochastic nature of the VAE means we can't be sure which one.

This raises a challenge for concept discovery, and more specifically for finding meets, since so far we've been assuming that the instances are provided to us deterministically. But in the probabilistic setting we're given two (or more generally a set of) inputs, and we'd like to find some concepts which apply to these inputs, by finding meets. However, we don't know for sure what the underlying instances are, since we only have a posterior over the instance space for each input. Hence we need a procedure for finding least general concepts which apply to some inputs with high probability, what we might call \emph{probabilistic meets}, and a way of propagating probabilities through hierarchical conceptual spaces.

The instance space now has a set of conditional probabilities associated with it (one for each input, $p(z\vert x)$), and so each instance now comes with some uncertainty associated with it.\footnote{We are following the standard convention of graphical models in using $x$ to denote the input and $z$ the hidden variable, which means that an instance is now denoted $z$, and the instance space $\mathcal{Z}$, rather than $x$ as before.}
Since the covariance matrix of $p(z\vert x)$ is assumed to be diagonal, the feature dimensions of $z$ can be treated independently, given the input, for example when calculating the probability of a concept, or finding the most probable concept. So when discussing $p(C\vert x)$, for example, for some concept $C$, this can either mean the full concept (in which case the probability would be the product of the probabilities for each dimension), or just the probability of $C$ along one dimension (i.e. the probability of a property). Which option is being used should be clear from the context.

How are the probabilities over instances propagated into the conceptual space? If we interpret $p(C\vert x)$ as the probability that $C$ is true given the input (i.e. that $C$ applies to $x$), then probabilities of concepts are given by weighted sums over instances, with the probabilities increasing monotonically with increases in the size of a concept's extension. Let $x$ be an input, $z$ an instance, $C$ a concept, $\mathcal{Z}(C)$ the extension of concept $C$, then:

\begin{eqnarray}
     p(C \vert x) & = &\int_{z \in \mathcal{Z}}    p(z\vert x) p(C \vert z) \\
    & = & \int_{z \in \mathcal{Z}(C)} p(z \vert x).
\end{eqnarray}

\noindent
Equation (1) follows from the rules of probability and the fact that $C$ is independent of $x$ given $z$. (2) follows from (1) since all our conceptual spaces considered so far have been deterministic, in the sense that an instance $z$ is either in the extension of a concept $C$ or not, with probability 1 or 0. 

If we denote the universal concept by $\mathcal{U}$, then $p(\mathcal{U} \vert x) = p(\mathcal{U} \vert z) = 1$.
We also have that, for a conceptual space with a partial order $\sqsubseteq$, if concept $C_1 \sqsubset C_2$, then $p(C_1 \vert x) > p(C_2 \vert x)$ and $p(C_1 \vert z) > p(C_2 \vert z)$.

\subsubsection{Finding meets with posterior distributions}
\label{sec:concepts_posterior}

As noted above, so far we have been assuming that, for the purposes of concept discovery from instances, the instances are provided with certainty, and then finding the meet---which could form part of a larger concept discovery process---is a deterministic operation which returns the least general concept consistent with the input instances. But what if there is a posterior distribution over the instances? This section outlines a Bayesian framework for thinking about finding meets probabilistically, assuming the conceptual space where the properties are real-valued closed intervals (with a maximal length). 

\begin{figure}
\begin{center}
     \makebox[\textwidth]{\includegraphics[width=13cm]{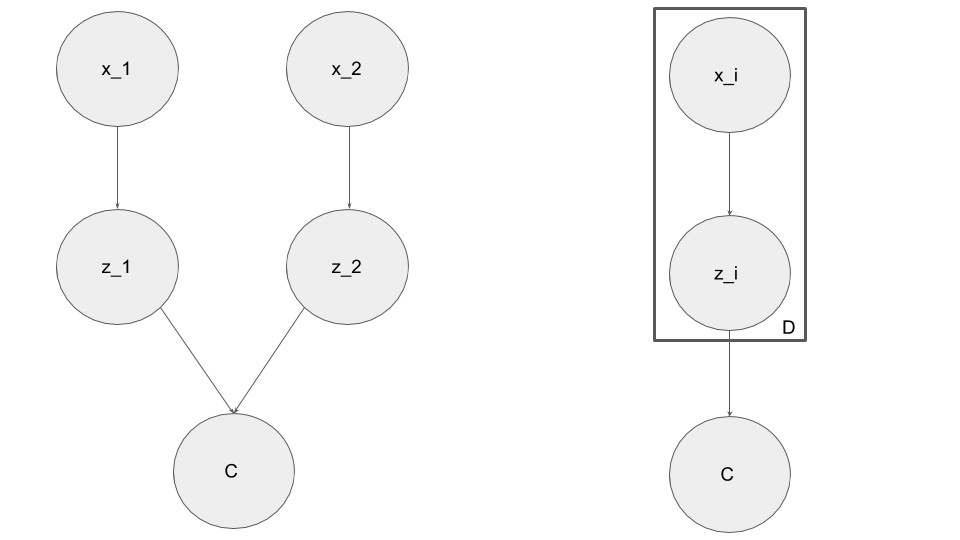}}
    \caption{Graphical model showing how $C$ relates to both inputs (on the left), and multiple inputs (on the right).}
    \vspace*{-0.5cm}
    \label{fig:graph_model}
    \end{center}
\end{figure}

Given two inputs $x_1, x_2$ (e.g. two images or videos), and an instance space $\mathcal{Z}$, then finding a probabilistic meet means informally finding a meet which is highly probable according to the posteriors $p(z_1\vert x_1)$ and $p(z_2\vert x_2)$. 
Let's continue with our interpretation of the $C$ random variable, where $C$ has the value $\mf{T}$ if concept $C$ holds, and $\mf{F}$ otherwise. As above, $C$ can be considered to refer to a single feature dimension, and have the value $\mf{T}$ if a particular property holds for that dimension. Now the goal is to, intuitively, find the least general property that has the highest probability of holding for both inputs. First of all, let's frame this problem as simply finding the most probable concept (i.e the concept that is most likely to apply to both inputs). Since the feature dimensions are conditionally independent (given the input), the optimisation can be carried out separately for each dimension. 

Let $\mathcal{C}$ be the conceptual space, $\mathcal{Z}(C)$ the extension of concept $C$, $\mathcal{C}_{\mbox{\scriptsize{max}}}$ the set of maximum-length property intervals in $\mathcal{C}$, and assume $C$ has the value $\mf{T}$ from (4) onwards, then:

\begin{eqnarray}
C_{\mbox{\scriptsize{opt}}} & = & \arg\max_{C\in\mathcal{C} \setminus \mathcal{U}}\; p(C = \mf{T}\vert x_1,x_2) \\
& = & \arg\max_{C\in\mathcal{C} \setminus \mathcal{U}}\int_{z_1,z_2} p(C,z_1,z_2\vert x_1,x_2) \\
& = & \arg\max_{C\in\mathcal{C} \setminus \mathcal{U}}\int_{z_1,z_2} p(z_1\vert x_1) p(z_2\vert x_2) p(C\vert z_1,z_2) \\
& = & \arg\max_{C\in\mathcal{C}_{\mbox{\scriptsize{max}}}}\int_{z_1,z_2} p(z_1\vert x_1) p(z_2\vert x_2) p(C\vert z_1,z_2) \\
& = & \arg\max_{C\in\mathcal{C}_{\mbox{\scriptsize{max}}}}\int_{z_1\in \mathcal{Z}(C)} p(z_1\vert x_1) \int_{z_2\in\mathcal{Z}(C)} p(z_2\vert x_2) 
\end{eqnarray}

First note that the maximisation is over $\mathcal{C} \setminus \mathcal{U}$, since the universal concept is always the most probable (with probability 1) and we don't necessarily want to return that. Since we're performing the optimisation for each feature dimension independently, $\mathcal{U}$ here is with respect to just one dimension, what we might call the \emph{universal property}. Line (4) follows from (3) by the rules of probability. (5) follows from (4) 
by the rules of probability and the conditional independence assumptions implied by Figure~\ref{fig:graph_model}. (6) follows from (5) since the probability of a concept being true increases monotonically with the concept extension's size, and $C_{\mbox{\scriptsize{max}}}$ is the set of concepts with the largest extensions (ignoring $\mathcal{U}$). And finally, (7) follows from (6) since the conditional probability of a concept being true is 0 outside of the concept's extension (hence the integals over $\mathcal{Z}(C)$) and is 1 within the extension (hence the dropping of $P(C\vert z_1,z_2)$).

\begin{figure}
\begin{center}
     \makebox[\textwidth]{\includegraphics[width=13cm]{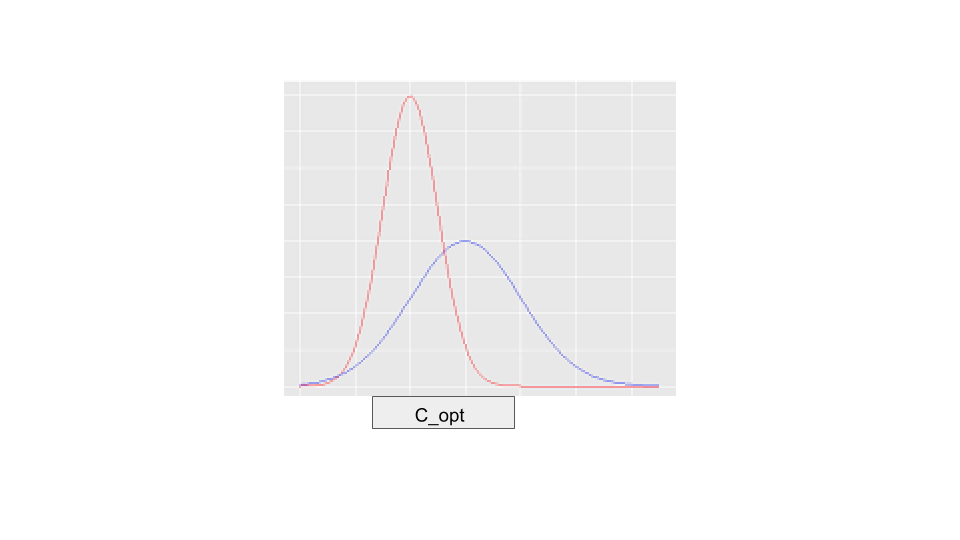}}
    \caption{Moving the maximum-length interval along the x-axis captures more or less of the probability mass from the two curves.}
    \vspace*{-0cm}
    \label{fig:overlap_gauss}
    \end{center}
\end{figure}

Figure~\ref{fig:overlap_gauss} gives some intuition behind the final equation above, showing how a maximum-length property interval can capture more or less of the probability mass from the two posteriors as it moves along the feature dimension.

One problem with this approach is that no abstraction will occur, since we are guaranteed a non-zero probability for each dimension (since the Gaussians corresponding to the two inputs will overlap to at least some degree on every dimension). An obvious solution is to only return a property for a dimension when the probability of that property holding for both inputs is above some threshold.

Another problem is that it always returns a maximum-length interval, whereas in practice we may want to return a smaller property. For example, consider a case where the two Gaussians are relatively peaked and the means close together; in this case we'd like to return a much smaller property interval since this will still contain most of the probability mass from both Gaussians. Here a natural solution would be to find the \emph{smallest} property interval \emph{with a probability above some threshold}. Since we're now searching for the smallest such interval---what we have called a ``probabilistic meet"---this is similar to finding the meet in the deterministic setting.\footnote{Whether there is guaranteed to be a unique smallest interval whose probability is above some threshold is left as an exercise for future work.} 

\begin{figure}
\begin{center}
     \makebox[\textwidth]{\includegraphics[width=13cm]{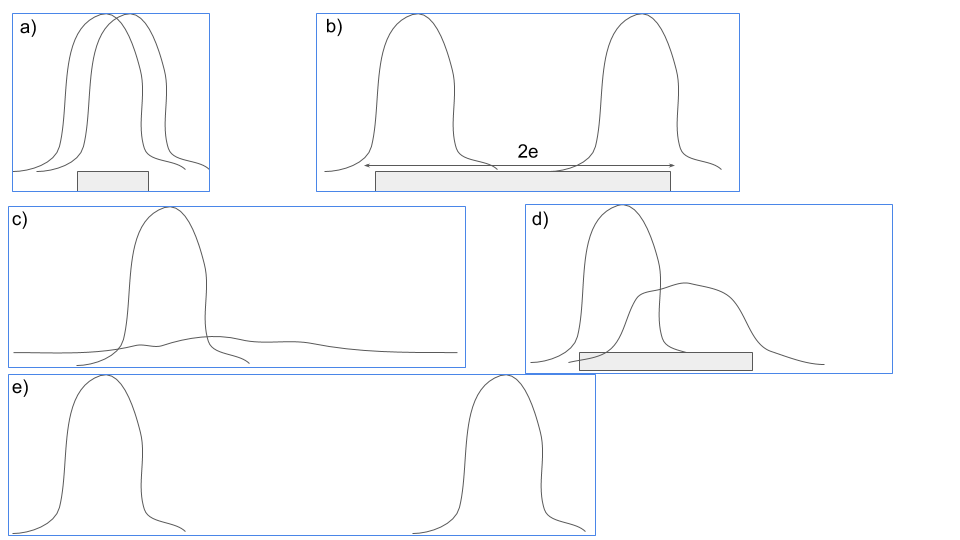}}
    \caption{Minimum-threshold probabilistic meets for different Gaussian pairs, where $2\epsilon$ is the maximum length for an interval.}
    \vspace*{-0cm}
    \label{fig:gaussians}
    \end{center}
\end{figure}

Figure~\ref{fig:gaussians} shows how a thresholded ``probabilistic meet" might look for different pairs of Gaussians, assuming a relatively high threshold (say 90\%). The maximum length for an interval is denoted 2$\epsilon$.
Example a) is a case where the means are similar and the variances are small, resulting in a short interval as the meet. Example b) has the two means relatively far apart, but still within the $2\epsilon$ maximum range and so the majority of the mass can be captured by a single interval. Example c) has both means close together, but one of the Gaussians has a large variance, and so there is no interval less then $2\epsilon$ which captures enough of the mass; in this case there is no meet. Example d) has one of the curves with a relatively large variance, but not so large that most of the mass cannot be captured in a single interval (since the means are also relatively close). And finally, e) is a case where the two variances are relatively small, but the means are too far apart to give a high-probability meet.  

A necessary extension is to consider the case where there are more than two inputs (the right side of Figure~\ref{fig:graph_model}). This is potentially non-trivial, however, since, for the properties-as-intervals conceptual space, the meet of a set of instances only depends on the outermost instances (because of the convexity condition). Hence this extension is left for future work. 

\begin{comment}
\scc{can we just assume, w.l.o.g., that the first two z's are always the outermost ones, in which case we can ignore the rest?}
\end{comment}

A further extension of the two approaches considered so far would be to introduce more stochasticity into the process by drawing $C$ probabilistically as well, given the two instances. That way it may be possible to have the most probable concept correspond to a reasonable probabilistic meet, but it would require a suitable prior over $C$, which we also leave for future work.

%% file: conclusion.tex
\section{Conclusion}

A key assumption in this work is that the representation learning algorithm provides separable feature dimensions (or more generally separable subspaces) so that whole dimensions can be dropped in the process of abstraction. Whether learning methods such as $\beta$-VAE have the appropriate biases to provide suitable feature dimensions is a question that requires substantial experimentation, continuing work such as that in \citeA{SCAN}. 

Furthermore, once we have the separable feature dimensions, then the process of concept discovery partly involves searching for sets of common properties across instances, which we have formalised as finding the meet of a set of instances. Continuing with our ongoing example, if a number of images from a platform video game all contain a heavy black ball, with the position and size of the ball varying, then removing the position and size features, but retaining weight, color and shape, could lead to the concept of a cannonball (assuming this particular feature set). However, it is likely that the weight, mass and color will vary slightly across these instances, and so it is not sufficient to search for identical real values. Hence for the real-valued feature spaces we have been considering, we had to insert additional structure into the feature lattices by grouping values into sets of reals, which then became the properties of the feature space. Whether a representation learning algorithm could induce this additional structure, as well as the separable feature dimensions, is also a question for future work.

Another place where our theorising in this report has raised interesting practical questions is in the interplay between conceptual spaces and probability distributions, especially the posteriors from a $\beta$-VAE. Section~\ref{sec:probs} set out some initial ideas in this direction, but there is still much to be done, especially in the potential role of probabilities in injecting some ``fuzziness" into the conceptual system. Note that the two conceptual spaces that have featured most prominently in this report---one which has discrete feature values from a partition of the reals, and one which has values as closed intervals---both have hard property boundaries, in the sense that two concepts can have values arbitrarily close along some feature dimension, but still have different property values for that feature. Whether this is a desirable feature of the space of base concepts, given the apparent ``fuzziness" of the human conceptual system \cite{laurence}, is open to debate.

The key mathematical insight from this work is encapsulated in Proposition~\ref{def:concept_space} from Section~\ref{sec:CPOs}, repeated here:

\begin{proposition}
(Repeat of prop.~\ref{def:concept_space}) A conceptual space of base concepts is a complete partial order (CPO) where the maximal elements of the CPO are instances in representation space, and the bottom element of the CPO is the universal concept.
\end{proposition}

A CPO beautifully captures the idea that instances are maximally informative concepts, with the least informative concept---the Universal Concept---sitting at the bottom of the CPO. It also neatly formalises the way in which abstraction over instances---by dropping whole feature dimensions---leads to concepts which lie below the relevant instances in the partial order, with the instances above a concept forming the concept's extension. Grouping of real values into property sets can also be easily accommodated in the partial order. A further advantage of the CPO is that the lattice structure comes with meets and joins which provide a natural mechanism for base concept combination. In practice, more complex mechanisms will be required for concept discovery and creation beyond base concepts. The hope is that this report has provided a sound theoretical basis on which to carry out such work.